 \documentclass[twocolumn,twoside]{IEEEtran}
\usepackage{amsmath,amssymb,amsthm,wasysym,epsfig,color,subfigure,graphicx,epstopdf,url,bm,nicefrac,tcolorbox}
\usepackage{caption}
\usepackage{amsmath,graphicx,bm,accents}
\usepackage{algorithm}
\usepackage{algorithmic}
\allowdisplaybreaks

\usepackage{hyperref}
\hypersetup{colorlinks=false}

\usepackage{accents}
\makeatletter
\def\wid{\check{{\cc@style\underline{\mskip9.5mu}}}}
\def\Wideubar{\underaccent{{\cc@style\underline{\mskip6mu}}}}
\makeatother

\makeatletter
\def\wideubar{\underaccent{{\cc@style\underline{\mskip9.5mu}}}}
\def\Wideubar{\underaccent{{\cc@style\underline{\mskip6mu}}}}
\makeatother

\makeatletter
\def\widebar{\accentset{{\cc@style\underline{\mskip9.5mu}}}}
\def\Widebar{\accentset{{\cc@style\underline{\mskip6mu}}}}
\makeatother

\newtheorem{proposition}{Proposition}

\newtheorem{theorem}{Theorem}

\theoremstyle{remark}\newtheorem{remark}{Remark}

\begin{document}
\title{Canonical Correlation Analysis of Datasets\\ with a Common Source Graph}

\author{Jia Chen,
	Gang Wang,~\IEEEmembership{Student Member,~IEEE},\\
	Yanning Shen,~\IEEEmembership{Student Member,~IEEE},
	and 
	Georgios B. Giannakis,~\IEEEmembership{Fellow,~IEEE}
	\thanks{This work was supported in part by NSF grants 1500713 and 1514056. 
	The authors are with the Digital Technology Center and the Department of Electrical and Computer Engineering, University of Minnesota, Minneapolis, MN 55455, USA. 
Emails: \{chen5625,\, gangwang, \,shenx513, \,georgios\}@umn.edu.
		}
}

\maketitle

\allowdisplaybreaks

\begin{abstract}
Canonical correlation analysis (CCA) is a powerful technique for discovering whether or not hidden sources are commonly present in two (or more) datasets. Its well-appreciated merits include dimensionality reduction, clustering,  classification, feature selection, and data fusion. The standard CCA however, does not exploit the geometry of the common sources, which may be available from the given data or can be deduced from (cross-) correlations. In this paper, this extra information provided by the common sources generating the data is encoded in a graph, and is invoked as a graph regularizer. This leads to a novel graph-regularized CCA approach, that is termed graph (g) CCA. The novel gCCA accounts for the graph-induced knowledge of common sources, while minimizing the distance between the wanted canonical variables. Tailored for diverse practical settings where the number of data is smaller than the data vector dimensions, the dual formulation of gCCA is also developed. One such setting includes kernels that are incorporated to account for nonlinear data dependencies. The resultant graph-kernel (gk) CCA is also obtained in closed form.  Finally, corroborating image classification tests over several real datasets are presented to showcase the merits of the novel linear, dual, and kernel approaches relative to competing alternatives.	
\end{abstract}


\section{Introduction}
\label{sec:intro}
In many fields, exploratory data analysis depends critically on dimensionality reduction, a process to discover compact representations of large volumes of high-dimensional data \cite{lle}.
Dimensionality reduction has been a crucial first step to obtain  tractable learning tasks, such as classification, clustering, and regression \cite{2002pca, lle}. 
Principal component analysis (PCA) is arguably the most widely used dimensionality reduction method, finding low-dimensional representations from high-dimensional data while preserving most of the data variance \cite{1901pca}.
Yet, ordinary PCA  presumes that data vectors lie close to a hyperplane - a gross geometrical approximation for several datasets. 
Local linear embedding on the other hand, preserves linear relationships between neighboring data \cite{lle}, while Laplacian eigenmaps ensure that data close in  the original manifold are mapped to close by locations in the low-dimensional space, thus aiming to preserve  local distances \cite{2003eigenmap}.

Nonetheless, such dimensionality reduction methods deal  with one dataset at a time. They
are challenged when it comes to analyzing two (or more) datasets jointly. Moreover, they require all data vectors to have the same dimension. Canonical correlation analysis (CCA) is a well-known method for extracting low-dimensional representations from two 
 datasets that can have  different dimensions, while maximizing their 
 correlations \cite{1936cca}. 
Although recent PCA variants such as discriminative PCA can deal with two datasets at a time, their goal is to extract the most discriminative features from the data of interest relative to the other \cite{dpca}. 
Formally, CCA aims at finding latent low-dimensional common structure from a paired dataset collected from different views of the same entities, also known as common sources.
  Each view contains high-dimensional representations of the sources in a certain feature space. For example, images of an individual captured by two cameras  can be interpreted as two different views of this individual (here playing the role of a source). 
  The ability of CCA to handle multiple datasets of different dimensions is a key enabler in tasks such as multi-mode data fusion,
where the need arises to fuse  information from
different domains \cite{hardoon2004canonical}. Ever since its proposition 
\cite{1936cca}, CCA benefits have been documented in diverse applications, 
 such as 
  blind source separation,  
brain imaging, 
clustering and classification, 
word embedding,  
and natural language processing, to name a few \cite{hardoon2004canonical,spm2010cca,tsp2013cca}.

To account for nonlinearities present in the data, kernel and deep CCA generalizations have also been  developed based on
 kernels or deep neural networks \cite{hardoon2004canonical,andrew2013deep}. Sparse CCA looking for sparse canonical vectors  
 was investigated by \cite{witten2009penalized}. Multi-view CCA on the other hand, generalizes  ordinary CCA to handle data from more than two modalities. Even though CCA solutions can be found via generalized eigen-decomposition, the resultant computational complexity may not scale well with the problem dimensionality. This motivated decentralized  CCA alternatives \cite{tsp2016cs}.

However, all aforementioned PCA and CCA tools do not exploit structural graph-induced information on the sources that may be available. 
Such information may be inferred from alternative views of the data, or it can be provided by the physics that dictates the underlying graph.
Indeed, graph-aware dimensionality reduction methods have lately demonstrated promising performance  \cite{gpca1, gpca,jstsp2016shahid,2012gdmf,spm2013emerging}.

Building on recent advances in  graph-aware dimensionality reduction \cite{gpca1,gpca}, 
the present paper introduces a neat link between graph embedding and canonical correlations, by putting forward a novel graph (g) CCA approach. Our gCCA pursues maximally correlated linear projections, while also leveraging statistical dependencies due
to the common sources hidden in the paired dataset. The underlying source  graph encoding these dependencies  can be either given, or be constructed based on prior knowledge.
When the number of data samples is smaller than the data vector dimensions, we advocate the graph dual (gd) CCA. Relative to gCCA, our gdCCA not only bypasses the inversion of  ill-conditioned data covariance matrices, but also incurs lower complexity in  high-dimensional setups. To further account for nonlinearities, we also develop what we term graph kernel (gk) CCA. Interestingly, solutions to all three gCCA variants can be found analytically through generalized eigenvalue decompositions. 

Different from \cite{blaschko2011semi, pr2014sun}, where CCA was regularized by two graph Laplacians separately per view, gCCA here jointly  leverages a single graph induced by the common sources. This is of major
practical importance, e.g., in brain mapping, where besides  functional magnetic resonance imaging (MRI) and diffusion-weighted MRI data collected at different brain regions \cite{brown2012ucla}, one has also access to the connectivity patterns among these regions. Finally,  numerical tests on several real-world datasets are presented to corroborate the merits of our proposed approaches for classification tasks over their competing alternatives.

The rest of this paper is structured as follows. Upon introducing the standard CCA in Section \ref{sec:review}, our gCCA is motivated, and derived in Section \ref{sec:gcca}. Its dual counterpart is developed in 
Section \ref{sec:gdcca}. Generalizing linear gCCA variants, the kernel version of gCCA is devised in Section \ref{sec:gkcca}. Numerical tests on several real-world datasets are presented in Section \ref{sec:test}, and the paper is concluded in Section \ref{sec:concl}. 

\emph{Notation}: Bold uppercase (lowercase) letters denote matrices (column vectors). Operators ${\rm Tr}(\cdot)$, $(\cdot)^{-1}$ and $(\cdot)^{\top}$ are matrix trace, inverse and transpose, respectively; $\|\cdot\|_2$ stands for the $\ell_2$-norm of vectors; $\mathbf{0}$ is an all-zero vector whose dimension is  clear from  the context; $ \langle \mathbf{a},\, \mathbf{b}\rangle$ denotes the inner product of vectors $\mathbf{a}$ and $\mathbf{b}$; and $\mathbf{I}$ represents the identity matrix of suitable size.

\section{Preliminaries}
\label{sec:review}
Consider  two datasets $\{\mathbf{x}_i\}_{i=1}^N$ and $\{\mathbf{y}_i\}_{i=1}^N$ with corresponding dimensionality $D_x$ and $D_y$,  collected from two different views of the same sources $\mathbf{s}_i\in\mathbb{R}^\rho $ with possibly $\rho \ll \min\{D_x,\,D_y\}$. CCA 
amounts to finding  low-dimensional subspaces  $\mathbf{U}\in\mathbb{R}^{D_x\times d}$ and $\mathbf{V}\in\mathbb{R}^{D_y\times d}$ with $d \leq  \rho$,  such that the Euclidean distance between the low-dimensional representations $\{\mathbf{U}^\top\mathbf{x}_i\}$ and $\{\mathbf{V}^\top\mathbf{y}_i\}$
is minimized. Assume without loss of generality that both datasets are centered, meaning  their corresponding sample means have been removed from the datasets. For ease of exposition, this section focuses on $d=1$ first, while generalization to $d\ge 2$ will be discussed later. CCA solves the following problem
\begin{subequations}
	\label{eq:occa}
\begin{align}	
\label{eq:cca_minc}
(\mathbf{u}^\ast,\,\mathbf{v}^\ast):=\arg\min_{\mathbf{u}, \,\mathbf{v}} ~&~~ \frac{1}{N}\sum_{i=1}^{N}\left(\mathbf{u}^\top \mathbf{x}_i - \mathbf{v}^{\top} \mathbf{y}_i\right)^2
\end{align}
where $\mathbf{u} \in \mathbb{R}^{D_x}$ and $\mathbf{v} \in \mathbb{R}^{D_y}$ are also termed a canonical pair. To ensure unique nonzero solutions however, the ensuing standard constraints are imposed 
\begin{align}\label{eq:ccaconst}
\mathbf{u}^\top \bm{\Sigma}_x \mathbf{u}  = 1,
\quad {\rm and}\quad 
\mathbf{v}^\top \bm{\Sigma}_y \mathbf{v}= 1
\end{align}
\end{subequations}
where $\bm{\Sigma}_x\! :=\! (1/N)\! \sum_{i=1}^N\mathbf{x}_i\mathbf{x}_i^{\top}$ and $\bm{\Sigma}_y\! :=\! (1/N)\sum_{i=1}^N\mathbf{y}_i\mathbf{y}_i^{\top}$ denote the sample covariance matrices of $\{\mathbf{x}_i\}$ and $\{\mathbf{y}_i \}$, respectively. 
Projections  $\{\mathbf{x}_i^{\top}\mathbf{u}^\ast\}_{i=1}^N$ and $\{\mathbf{y}_i^{\top}\mathbf{v}^\ast\}_{i=1}^N$ form  a pair of canonical variables,
which can be interpreted as low-dimensional approximations of the common sources $\{\mathbf{s}_i \}_{i=1}^N$. 

After simple manipulations, \eqref{eq:occa} leads to the following popular formulation of CCA \cite{hardoon2004canonical} 
\begin{subequations}
	\label{eq:cca}
	\begin{align}
	(\mathbf{u}^\ast, \mathbf{v}^\ast):=\arg\max_{\mathbf{u}, \,\mathbf{v}} ~&~ \mathbf{u}^\top \bm{\Sigma}_{xy} \mathbf{v}\label{eq:ccac}\\
	\rm{s.\, to} ~&~  \mathbf{u}^\top \bm{\Sigma}_x \mathbf{u} = 1,~ {\rm and}~\mathbf{v}^\top \bm{\Sigma}_y \mathbf{v} = 1\label{eq:ccac2}
	\end{align}
\end{subequations}
where $\bm{\Sigma}_{xy}:=(1/N)\sum_{i=1}^N\mathbf{x}_i\mathbf{y}_i^{\top}$ is  the sample cross-covariance matrix of $\{\mathbf{x}_i\}$ and $\{\mathbf{y}_i \}$.

Using Lagrange duality theory, 
the solution of \eqref{eq:cca} will  be given next  in analytical form. To this end, letting $\lambda,\,\mu\in\mathbb{R}$ be the dual variables associated with the two constraints in \eqref{eq:ccac2}, one can write the Lagrangian as 
\begin{equation*}
\mathcal{L}(\mathbf{u},\mathbf{v};\lambda,\mu)=\mathbf{u}^\top\bm{\Sigma}_{xy}\mathbf{v}-\!\lambda(\mathbf{u}^\top \bm{\Sigma}_x \mathbf{u} -1)-\mu(\mathbf{v}^\top \bm{\Sigma}_y \mathbf{v} - 1).
\end{equation*} 
At the optimum  $	(\mathbf{u}^\ast, \,\mathbf{v}^\ast)$, the KKT conditions assert that
\begin{subequations}
	\begin{align}
	&\bm{\Sigma}_{xy}\mathbf{v}^\ast=2\lambda^\ast \bm{\Sigma}_x\bm{u}^\ast,\qquad(\mathbf{u}^\ast)^\top \bm{\Sigma}_x \mathbf{u}^\ast = 1\label{eq:ccalag1}\\
	&\bm{\Sigma}_{xy}^\top\mathbf{u}^\ast=2\mu^\ast \bm{\Sigma}_y\mathbf{v}^\ast,\qquad (\mathbf{v}^\ast)^\top \bm{\Sigma}_y \mathbf{v}^\ast = 1.\label{eq:ccalag2}
	\end{align}
\end{subequations}

Left-multiplying the first equations in \eqref{eq:ccalag1} and \eqref{eq:ccalag2} by $(\mathbf{u}^\ast)^\top$ and $(\mathbf{v}^\ast)^\top$, respectively, lead  to $(\mathbf{u}^\ast)^\top\bm{\Sigma}_{xy}\mathbf{v}^\ast=2\lambda^\ast=2\mu^\ast$. Hence, solving \eqref{eq:cca} reduces to solving the generalized eigenvalue problem, see e.g.,  \cite{hardoon2004canonical}
\begin{equation}
\label{eq:ccasol}
\left[\begin{array}
{cc}
\bm{\Sigma}_{xy}^\top&\mathbf{0}\\
\mathbf{0}&\bm{\Sigma}_{xy}
\end{array}
\right]\left[\begin{array}
{c}
\mathbf{u}\\
\mathbf{v}
\end{array}
\right]=2\lambda\left[\begin{array}
{cc}
\mathbf{0}&\bm{\Sigma}_{y}\\
\bm{\Sigma}_{x}&	\mathbf{0}
\end{array}
\right]\left[\begin{array}
{c}
\mathbf{u}\\
\mathbf{v}
\end{array}
\right].
\end{equation}
Maximizing the objective function \eqref{eq:ccac} is tantamount to finding  the largest generalized eigenvalue $\lambda^\ast:=\lambda_1$ in \eqref{eq:ccasol}, and the optimal canonical vectors $[(\mathbf{u}^\ast)^\top~(\mathbf{v}^\ast)^\top]^\top$ to \eqref{eq:cca} are obtained from  the corresponding generalized eigenvector. 

In order to find $d\le \min (D_x,\, D_y)$ pairs of canonical vectors, say $\{(\mathbf{u}_i, \mathbf{v}_i)\}_{i=1}^d$, one can basically repeat the steps leading to (5) with extra constraints. Specifically, if the first $(k-1)$ pairs  $\{(\mathbf{u}_i^\ast,\,\mathbf{v}_i^\ast)\}_{i=1}^{k-1}$ have been found, 
the $k$-th pair can be obtained by solving \eqref{eq:cca} with the orthogonality constraints $(\mathbf{u}^\ast_k)^\top \bm{\Sigma}_x \mathbf{u}_i^\ast = 0$ and $(\mathbf{v}_k^\ast)^\top \bm{\Sigma}_y \mathbf{v}_i^\ast = 0$ for $i=1,\,2,\,\ldots,\, k-1$; that is,
\begin{subequations}
	\label{eq:ccarecu}
	\begin{align}
	\max_{\mathbf{u}_k, \,\mathbf{v}_k} ~&~~ \mathbf{u}_k^\top \bm{\Sigma}_{xy} \mathbf{v}_k\label{eq:ccarecuc}\\
	{\rm s. \,to} ~&~~  \mathbf{u}_k^\top \bm{\Sigma}_x \mathbf{u}_k = 1,\quad
	\mathbf{v}_k^\top \bm{\Sigma}_y \mathbf{v}_k = 1
 \label{eq:ccarecuc1}\\
	&~~	\mathbf{u}_k^\top \bm{\Sigma}_x \mathbf{u}_i^\ast = 0,\quad \mathbf{v}_k^\top \bm{\Sigma}_y \mathbf{v}_i^\ast = 0\label{eq:ccarecuc1}\\
	&~~\forall i = 1,\,2,\,\ldots,\,k-1
	\end{align}
\end{subequations}
and the same steps  can be repeated until $d$ canonical pairs are found.
For brevity, let us concatenate the $d$ canonical vectors $\{\mathbf{u}_i\}$ and $\{\mathbf{v}_i\}$ to form matrices $\mathbf{U}$ and $\mathbf{V}$ accordingly, and rewrite \eqref{eq:ccarecu} in the following compact form
\begin{subequations}\label{eq:ccam}
\begin{align}
\max_{\mathbf{U}, \, \mathbf{V}} ~&~~{\rm Tr}(\mathbf{U}^\top \bm{\Sigma}_{xy} \mathbf{V})\\
{\rm s.\,to}~&~~\mathbf{U}^\top \bm{\Sigma}_x \mathbf{U} = \mathbf{I},\quad  {\rm and}\quad\mathbf{V}^\top \boldsymbol{\Sigma}_y \mathbf{V} = \mathbf{I}
\end{align}
\end{subequations}
which yields simultaneously multiple canonical vectors. As deduced earlier, the $m$-th columns of minimizers $\mathbf{U}^\ast\in\mathbb{R}^{D_x\times d}$ and  $\mathbf{V}^\ast\in\mathbb{R}^{D_y\times d}$ of \eqref{eq:ccam}  
 correspond to the left and right generalized eigenvectors of \eqref{eq:ccasol} associated with the $m$-th largest generalized eigenvalue, respectively.

\section{CCA over Graphs}
\label{sec:gcca}
In diverse applications, the common sources $\{\mathbf{s}_i\}_{i=1}^N$ may be viewed as nodal vectors of a graph having $N$ nodes. This  structural prior information can be  leveraged when finding  the canonical vectors. 
In this paper, this extra knowledge of common sources is encoded in a graph, and will be  embodied in the canonical variables through graph regularization.

We outline some basics of the graph theory first. 
A graph is represented by a tuple $\mathcal{G}=\{\mathcal{N},\,\mathcal{W}\}$, where $\mathcal{N}:=\{1,\,2,\,\ldots,\,N\}$ is the vertex set, and $\mathcal{W}:=\{w_{ij}\}_{(i,j)\in \mathcal{N}\times\mathcal{N}} $ stacks up edge  weights $w_{ij}$ over all vertex pairs $(i,\,j)$. For ease of exposition, this paper focuses on undirected graphs, for which $w_{ij}=w_{ji}$ for all $i,\,j\in\mathcal{N}$. Moreover, a graph is said to be unweighted if all $w_{ij}$'s take binary values $0$ or $1$.  
Upon forming the so-called weighted adjacency matrix $\mathbf{W}\in\mathbb{R}^{N\times N}$ with its $(i,j)$-th entry being $w_{ij}$, and defining $d_{i}:=\sum_{j=1}^N {w}_{ij}$, the Laplacian matrix of graph $\mathcal{G}$ is given by  
\begin{equation}
\label{eq:lg}
\mathbf{L}_{\mathcal{G}} := \mathbf{D} - \mathbf{W} \in \mathbb{R}^{N \times N}
\end{equation}
where the diagonal matrix $\mathbf{D} \in \mathbb{R}^{N \times N}$ holds ordered  entries  $\{d_{i}\}_{i=1}^N$ on its diagonal. 

Having introduced basic graph notation,  we present a neat link between canonical correlations and graph embedding next. 
Consider for instance a graph $\mathcal{G}$ with adjacency matrix $\mathbf{W}$, over which the underlying sources $\{\mathbf{s}_i\}_{i=1}^N$ are assumed to be smooth. In other words, vectors $(\mathbf{s}_i,\, \mathbf{s}_j)$ residing on two connected nodes  $i,\,j\in \mathcal{G}$ are deemed close to each other in Euclidean distance. As remarked earlier, canonical variables $\mathbf{u}^\top\mathbf{x}_i$ and $\mathbf{v}^\top\mathbf{y}_j$ are accordingly one-dimensional approximates of $\mathbf{s}_i$ and $\mathbf{s}_j$. Building on this fact, let us now focus on the weighted sum of distances between any two pairs of canonical variables from $\{\mathbf{u}^\top\mathbf{x}_i\}_{i=1}^N$ and $\{\mathbf{v}^\top\mathbf{y}_i\}_{i=1}^N$ over $\mathcal{G}$, namely the  quadratic term
\begin{equation}
\label{eq:regg}
\sum_{i=1}^N \sum_{j=1}^N{w}_{ij}\!\left(\mathbf{u}^\top\mathbf{x}_i - \mathbf{v}^\top\mathbf{y}_j\right)^2.
\end{equation}

It is clear that by minimizing \eqref{eq:regg} over $\mathbf{u}$ and $\mathbf{v}$, canonical variables $\mathbf{u}^\top\mathbf{x}_i$ and  $\mathbf{v}^\top\mathbf{y}_j$ corresponding to adjacent nodes $i,\,j\in\mathcal{G}$ with large edge weights $w_{ij}$ will be promoted to stay close to each other.
As such, invoking this term as a regularizer
 accounts for the additional graph
knowledge of the common sources, while maximizing the linear
correlation coefficient between the canonical variables, yielding
\begin{align*}
\min_{\mathbf{u},\,\mathbf{v}} ~&\,
\frac{1}{2N}\!\sum_{i=1}^N\!\left(\mathbf{u}^{\top}\!\mathbf{x}_i\! -\! \mathbf{v}^{\top}\!\mathbf{y}_i\right)^2\!
+ \! \frac{\gamma}{2}\!\sum_{i=1}^N\! \sum_{j=1}^N{w}_{ij}\!\left(\mathbf{u}^\top\!\mathbf{x}_i -\! \mathbf{v}^\top\!\mathbf{y}_j\right)^2\nonumber\\
{\rm s.\,to}~&\, \mathbf{u}^{\top}\boldsymbol{\Sigma}_x\mathbf{u} = 1,\quad {\rm and}\quad  \mathbf{v}^{\top}\boldsymbol{\Sigma}_y\mathbf{v} = 1
\end{align*}
in which $\gamma\ge 0$ is a hyper-parameter that balances the distance between canonical variable estimates with their smoothness over $\mathcal{G}$. After expanding the squares and removing the constant terms, the problem at hand can be equivalently rewritten as 
\begin{subequations}
	\label{eq:gcca_o}
	\begin{align}
	\max_{\mathbf{u},\,\mathbf{v}}~&~
	\mathbf{u}^{\top}\bm{\Sigma}_{xy}\mathbf{v}\!-\!\gamma\mathbf{u}^{\top}\mathbf{X}\mathbf{L}_{\mathcal{G}}\mathbf{Y}^{\top}\mathbf{v}
\!	-\!\frac{\gamma}{2}\sum_{i=1}^N d_i\!\left(\mathbf{u}^{\top}\mathbf{x}_i\!-\!\mathbf{v}^{\top}\mathbf{y}_i\right)^2\label{eq:gcca_oc}\\
	{\rm s.\,to}~&~ \mathbf{u}^{\top}\boldsymbol{\Sigma}_x\mathbf{u} = 1,\quad {\rm and}\quad \mathbf{v}^{\top}\boldsymbol{\Sigma}_y\mathbf{v} = 1.\label{eq:gcca_oc2}
	\end{align}
\end{subequations}

Evidently, problem \eqref{eq:gcca_o} is non-convex and is not amenable to efficient solvers due to the bilinear terms as well as the quadratic equality constraints. Even though block coordinate descent-type solvers can be employed, only convergence to a stationary point can be guaranteed in general \cite{tsp2016cs}. Instead of coping with the objective function \eqref{eq:gcca_oc} directly, we shall pursue a lower bound of it, which will turn out to afford  an analytical solution.

Toward that end, it is easy to verify that with all $\{d_i\ge 0\}_{i=1}^N$, the following holds for all $\mathbf{u}\in\mathbb{R}^{D_x}$ and $\mathbf{v}\in\mathbb{R}^{D_y}$:
\begin{equation}
\sum_{i=1}^N d_i\!\left(\mathbf{u}^{\top}\mathbf{x}_i-\mathbf{v}^{\top}\mathbf{y}_i\right)^2\le 
2 d_{\max}N\left(\mathbf{u}^{\top}\boldsymbol{\Sigma}_x\mathbf{u}+\mathbf{v}^{\top}\boldsymbol{\Sigma}_y\mathbf{v} \right)\label{eq:lbound}
\end{equation}
where $d_{\max}:=\max_{1\le i\le N} d_i$, and the equality is achieved when $d_i=d_{\max}$ and $\mathbf{u}^\top\mathbf{x}_i=-\mathbf{v}^\top\mathbf{y}_i$ for all $i=1,\,2,\,\ldots,\,N$. 
Subsequently, we replace the last term in \eqref{eq:gcca_oc} with the right-hand-side term, which contributes to a valid lower bound of \eqref{eq:gcca_oc}. Formally stated, we have the following reformulation.

	\begin{proposition}
		\label{prop:gcca}
		Replacing the sum in \eqref{eq:gcca_oc} with its upper bound in \eqref{eq:lbound}  
		leads to an objective that lower bounds \eqref{eq:gcca_oc}. Merging and ignoring the constant terms due to the equality constraints \eqref{eq:gcca_oc2} leads to our novel gCCA formulation 
			\end{proposition}
			\begin{tcolorbox}\vspace{-8pt}
		\begin{subequations}
			\label{eq:gccav}
			\begin{align}
			\max_{\mathbf{u},\,\mathbf{v}}~&~~\mathbf{u}^{\top}\bm{\Sigma}_{xy}\mathbf{v}-\gamma\mathbf{u}^{\top}\mathbf{X}\mathbf{L}_{\mathcal{G}}\mathbf{Y}^{\top}\mathbf{v}\\
			\rm{s.\,to} ~&~~  \mathbf{u}^\top \bm{\Sigma}_x \mathbf{u} = 1,\quad {\rm and}\quad \mathbf{v}^\top \bm{\Sigma}_y \mathbf{v} = 1.
			\end{align}
		\end{subequations}
\end{tcolorbox}

Clearly, when $\gamma=0$, our gCCA finds $(\mathbf{u}, \,\mathbf{v})$ that only maximizes the linear correlation between the pair of canonical variables. In this case, no graph knowledge is exploited, and our gCCA reduces to the standard CCA. With $\gamma$ increasing gradually, gCCA accounts  progressively for  extra graph information of the common sources when finding  the canonical variables.

Next, let us consider multiple canonical pairs $\{(\mathbf{u}_i,\,\mathbf{v}_i) \}_{i=1}^d$, and collect them to form matrices $\mathbf{U}:=[\mathbf{u}_1~\cdots~\mathbf{u}_d ]$ and $\mathbf{V}:=[\mathbf{v}_1~\cdots~\mathbf{v}_d] $. We can then generalize gCCA in \eqref{eq:gccav} to $d\ge 2$ as 
\begin{subequations}
	\label{eq:gpcca}
	\begin{align}
	\max_{\mathbf{U}, \,\mathbf{V}}~&~~{\rm Tr}\!\left(\mathbf{U}^{\top}\bm{\Sigma}_{xy}\mathbf{V}-\gamma \mathbf{U}^{\top}\mathbf{X}\mathbf{L}_{\mathcal{G}}\mathbf{Y}^{\top}\mathbf{V}\right)\label{eq:gpccac}\\
	{\rm s.\, to }~&~~\mathbf{U}^{\top}\bm{\Sigma}_{x}\mathbf{U}=\mathbf{I},\quad {\rm and}\quad \mathbf{V}^{\top}\bm{\Sigma}_{y}\mathbf{V}=\mathbf{I}.\label{eq:gpccac2}
	\end{align}
\end{subequations}
Interestingly, even with the  extra graph-inducing regularization term, our gCCA in \eqref{eq:gpcca} still admits an analytical solution, under the standard assumption that data covariance matrices $\bm{\Sigma}_x$ and $\bm{\Sigma}_y$ are both nonsingular. For concreteness, the solution is summarized in the following result, and for self-contained presentation, its proof is provided in Appendix \ref{prof2}. 

\begin{theorem}\label{thm:gcca}
	Given zero-mean data $\{\mathbf{x}_i\in\mathbb{R}^{D_x} \}_{i=1}^N$ and $\{\mathbf{y}_i\in\mathbb{R}^{D_y} \}_{i=1}^N$, suppose that $\bm{\Sigma}_x=(1/N)\!\sum_{i=1}^N\!\mathbf{x}_i\mathbf{x}_i^{\top}$ and $\bm{\Sigma}_y=(1/N)\sum_{i=1}^N\mathbf{y}_i\mathbf{y}_i^{\top}$ are nonsingular. Then the optimal solution  $(\mathbf{U}^\ast\in\mathbb{R}^{D_x\times d} ,\,\mathbf{V}^\ast\in\mathbb{R}^{D_y\times d})$ to the gCCA problem \eqref{eq:gpcca} with $d\le\min(D_x, D_y)$, is  given by
	\begin{equation}
	\label{eq:gpcca_sol}
	\mathbf{U}^\ast:=\bm{\Sigma}_x^{-1/2}\bar{\mathbf{U}}^\ast, \quad{\rm and}\quad  \mathbf{V}^\ast:=\bm{\Sigma}_y^{-1/2}\bar{\mathbf{V}}^\ast
	\end{equation}
	where the columns of $\bar{\mathbf{U}}^\ast\in\mathbb{R}^{D_x\times d}$ and $\bar{\mathbf{V}}^\ast\in\mathbb{R}^{D_y\times d}$ are the $d$ left and right singular vectors of $\bm{\Sigma}_x^{-1/2}\left(\bm{\Sigma}_{xy}-\gamma\mathbf{X}\mathbf{L}_{\mathcal{G}}\mathbf{Y}^{\top}\right)\bm{\Sigma}_y^{-1/2}$ associated with its $d$ largest singular values.
	Moreover, the maximum objective value of \eqref{eq:gpccac} is the sum  of the $d$ largest singular values.
\end{theorem}

Our  proposed gCCA scheme is summarized in Alg. \ref{alg:gcca}.  
Two remarks are now  in order. 
\begin{remark}
	Different from our single  regularizer in \eqref{eq:gpcca}, the approaches  in \cite{blaschko2011semi,pr2014sun} rely on two regularizers
 or two constraints involving graph priors $\mathbf{U}^{\top}\mathbf{X}\mathbf{L}_{\mathcal{G}_{ x}}\mathbf{X}^{\top}\mathbf{U}$ and $\mathbf{V}^{\top}\mathbf{Y}\mathbf{L}_{\mathcal{G}_{ y}}\mathbf{Y}^{\top}\mathbf{V}$ for the two-view data $\mathbf{X}$ and $\mathbf{Y}$, respectively. However, the problem formulation in \cite{pr2014sun} does not admit an analytical solution. Although iterative algorithms can be used to solve the involved nonconvex optimization problem, only convergence to a stationary point can be ensured in general \cite{bertsekas1999nonlinear}.   
 When the two datasets lie in two distinct graphs $\mathcal{G}_{ x}$ and $\mathcal{G}_{ y}$, using 
the graph-Laplacian regularized constraints can improve standard CCA performance \cite{belkin2006manifold}. This approach is mainly suggested for semi-supervised learning, where $\bm{\Sigma}_{xy}$ is fully available.
 In contrast, \eqref{eq:gpcca} leverages the graph induced by the common sources, and our source graph regularizer $\mathbf{U}^{\top}\mathbf{X}\mathbf{L}_{\mathcal{G}}\mathbf{Y}^{\top}\mathbf{V}$ directly exploits correlations between the low-dimensional approximations of common sources over  $\mathcal{G}$. 
	 This is critical in certain practical setups, in which one has prior knowledge about the common sources besides the given datasets. In brain imaging for instance, in addition to the functional MRI and diffusion-weighted MRI data collected at different brain regions \cite{brown2012ucla}, one has also access to the connectivity patterns among these regions. Furthermore, our proposed gCCA framework admits an analytical solution.
\end{remark} 
\begin{remark}\label{rmk:4}
	To induce different graph properties, rather than relying on $\mathbf{L}_{\mathcal{G}}$, a family of graph regularizations of the form $r(\mathbf{L}_{\mathcal{G}}):=\sum_{i=1}^N r(\lambda_{i}^{w})\mathbf{u}^w_i(\mathbf{u}^w_i)^{\top}$ can be also employed \cite{graphkernel}, where  $r(\cdot):\mathbb{R}\to\mathbb{R}^+$ is a scalar function, and appropriate choices of $r(\lambda^w_i)$ are helpful for inducing diverse graph properties; while  
	$\mathbf{u}^w_i\in\mathbb{R}^{N}$ is the eigenvector of $\mathbf{L}_{\mathcal{G}}$ associated with its $i$-th largest eigenvalue $\lambda^w_i$. 
\end{remark}

\begin{algorithm}[t]
	\caption{CCA with a common source graph\textcolor{red}{}}
	\label{alg:gcca}
	\begin{algorithmic}[1]
		\STATE {\bfseries Input:}
		$\{\mathbf{x}_i\}_{i=1}^N$, $\{\mathbf{y}_i\}_{i=1}^N$, $d$, $\mathbf{W}$, and $\gamma$.
		\STATE {\bfseries Form} (cross-)covariance matrices, $\bm{\Sigma}_x$, $\bm{\Sigma}_y$ and $\bm{\Sigma}_{xy}$.
		\STATE {\bfseries Build} $\mathbf{L}_{\mathcal{G}}$ using \eqref{eq:lg}.
		\STATE {\bfseries Perform} SVD on $\bm{\Sigma}_x^{-1/2}\left(\bm{\Sigma}_{xy}-\gamma\mathbf{X}\mathbf{L}_{\mathcal{G}}\mathbf{Y}^{\top}\right)\bm{\Sigma}_y^{-1/2}$ 
		\STATE{\bfseries Extract} the first $d$ leading eigenvectors to obtain  $\bar{\mathbf{U}}^\ast$ and $\bar{\mathbf{V}}^\ast $.
		\STATE{\bfseries Compute} $\mathbf{U}^\ast=\bm{\Sigma}_x^{-1/2}\bar{\mathbf{U}}^\ast$ and $\mathbf{V}^\ast=\bm{\Sigma}_y^{-1/2}\bar{\mathbf{V}}^\ast$.
		\STATE {\bfseries Output:}\label{step:5} $\mathbf{U}^\ast$ and  $\mathbf{V}^\ast$.
	\end{algorithmic}
\end{algorithm}

\section{Dual CCA over Graphs}
\label{sec:gdcca}

Similar to dual PCA \cite{gpca}, various practical scenarios involving high-dimensional data vectors, have  $N\ll \min \{D_x,\,D_y\}$, in which case $\bm{\Sigma}_x$ and $\bm{\Sigma}_y$ become singular, and the results in Theorem \ref{thm:gcca} do not apply. Even though this rank deficiency can be remedied with appropriate Tikhonov regularization \cite{hardoon2004canonical}, the resultant computational complexity can be considerably higher than the alternative of investigating gCCA in the dual domain. 
In this direction, consider first expressing  $\mathbf{u}\in\mathbb{R}^{D_x}$ and $\mathbf{v}\in\mathbb{R}^{D_y}$ in terms of their corresponding parts of the data matrices $\mathbf{X}$ and $\mathbf{Y}$ as
\begin{equation}
\label{eq:express}
\mathbf{u}:=\mathbf{X}\bm{\alpha},\quad{\rm and}\quad \mathbf{v}:=\mathbf{Y}\bm{\beta}
\end{equation} 
where $\bm{\alpha} \in \mathbb{R}^{N}$ and $\bm{\beta} \in \mathbb{R}^{N}$ are the so-termed dual vectors. 
Substituting \eqref{eq:express} into \eqref{eq:gccav} gives rise to our graph dual (gd) CCA formulation for one pair of canonical vectors
\begin{subequations}
	\label{eq:gcca}
	\begin{align}
	\max_{\bm{\alpha},\,\bm{\beta}}~&~~
	\bm{\alpha}^\top\mathbf{X}^\top\mathbf{X}\mathbf{Y}^\top \mathbf{Y}\bm{\beta} -
	\gamma	\bm{\alpha}^\top\mathbf{X}^\top\mathbf{X}\mathbf{L}_{\mathcal{G}} \mathbf{Y}^\top \mathbf{Y}\bm{\beta}\label{eq:gccac}\\
	\rm{s.\,to}~&~~
	\bm{\alpha}^\top \mathbf{X}^\top \mathbf{X} \mathbf{X}^\top \mathbf{X} \bm{\alpha} = 1\label{eq:gccac1}\\
	~&~~
	\bm{\beta}^\top \mathbf{Y}^\top \mathbf{Y} \mathbf{Y}^\top \mathbf{Y} \bm{\beta} = 1.\label{eq:gccac2}
	\end{align}
\end{subequations}

Similar  to Section \ref{sec:gcca}, introducing variables $\lambda_x\in\mathbb{R}$ and $\lambda_y\in\mathbb{R}$ to be the Lagrange multipliers corresponding to constraints \eqref{eq:gccac1} and \eqref{eq:gccac2}, respectively, one can write the Lagrangian for \eqref{eq:gcca} as
\begin{align} 
&\mathcal{L}(\bm{\alpha}, \,\bm{\beta};\,\lambda_x,\, \lambda_y) := -\bm{\alpha}^\top\mathbf{X}^\top\mathbf{X}(\mathbf{I} -{\gamma} \mathbf{L}_{\mathcal{G}}) \mathbf{Y}^\top \mathbf{Y}\bm{\beta}\nonumber \\
&+ \frac{\lambda_x}{2} (\bm{\alpha}^\top \mathbf{X}^\top \mathbf{X} \mathbf{X}^\top \mathbf{X} \bm{\alpha}-1) + \frac{\lambda_y}{2} (\bm{\beta}^\top \mathbf{Y}^\top \mathbf{Y} \mathbf{Y}^\top \mathbf{Y} \bm{\beta}-1).\nonumber
\end{align} 
Setting derivatives of the Lagrangian  with respect to $\bm{\alpha}$ and $\bm{\beta}$ to zero further leads to
\begin{subequations}
	\label{eq:dcca:derivative}	
	\begin{align}
	-\mathbf{X}^\top\mathbf{X}(\mathbf{I} -\gamma \mathbf{L}_{\mathcal{G}}) \mathbf{Y}^\top \mathbf{Y}\bm{\beta}
	+  \lambda_x \mathbf{X}^\top \mathbf{X} \mathbf{X}^\top \mathbf{X} \bm{\alpha} & = \mathbf{0}
	\label{eq:lag_deri1}\\
	-\mathbf{Y}^\top\mathbf{Y}(\mathbf{I} -\gamma \mathbf{L}_{\mathcal{G}}) \mathbf{X}^\top \mathbf{X}\bm{\alpha}+  \lambda_y  \mathbf{Y}^\top \mathbf{Y} \mathbf{Y}^\top \mathbf{Y} \bm{\beta} &= \mathbf{0}	\label{eq:lag_deri2}.
	\end{align}
\end{subequations}
Left-multiplying \eqref{eq:lag_deri1} and \eqref{eq:lag_deri2} by $\bm{\alpha}^{\top}$ and $\bm{\beta}^{\top}$, respectively, and subsequently subtracting the latter from the former, we arrive at
\begin{equation}
\label{eq:lag_subt}
\lambda_x \bm{\alpha}^{\top}\mathbf{X}^\top \mathbf{X} \mathbf{X}^\top \mathbf{X} \bm{\alpha} - \lambda_y \bm{\beta}^{\top} \mathbf{Y}^\top \mathbf{Y} \mathbf{Y}^\top \mathbf{Y} \bm{\beta} =0.
\end{equation}
Taking into account \eqref{eq:lag_subt}, \eqref{eq:gccac1}, and \eqref{eq:gccac2}, it follows  that at the optimal solution, we have $\lambda^\ast:=\lambda_x^\ast = \lambda_y^\ast$. 
Supposing  for now that $\bf{X}^{\top}\bf{X}$ and $\bf{Y}^{\top}\bf{Y}$ are nonsingular, we find 
\begin{equation}
\label{eq:lag_relation}
{\bm{\alpha}}^\ast := \frac{1}{\lambda^\ast} \left(\mathbf{X}^{\top}\mathbf{X}\right)^{-1}\left(\mathbf{Y}^{\top}\mathbf{Y} - \gamma\mathbf{L}_{\mathcal{G}}\mathbf{Y}^{\top}\mathbf{Y}\right)\bm{\beta}^{\ast}.
\end{equation} 
Plugging \eqref{eq:lag_relation} into \eqref{eq:lag_deri2} yields
\begin{equation}
\label{eq:lag_beta}
\left(\mathbf{Y}^{\top}\mathbf{Y}\right)^{-1}\left(\mathbf{I}-{\gamma}\mathbf{L}_{\mathcal{G}}\right)^2\mathbf{Y}^{\top}\mathbf{Y}\bm{\beta}^{\ast} = (\lambda^\ast)^2 \bm{\beta}^{\ast}
\end{equation}
and similarly, one obtains that
\begin{equation}
\label{eq:lag_alpha}        
\left(\mathbf{X}^{\top}\mathbf{X}\right)^{-1}\left(\mathbf{I}-{\gamma}\mathbf{L}_{\mathcal{G}}\right)^2\mathbf{X}^{\top}\mathbf{X}\bm{\alpha}^{\ast} = (\lambda^\ast)^2 \bm{\alpha}^{\ast}.
\end{equation}
The last two equalities show that $\bm{\alpha}^\ast$ depends solely on $\mathbf{X}$, and $\bm{\beta}^\ast$ solely on $\mathbf{Y}$. This holds without any assumption about the paired dataset $\mathbf{X}$ and $\mathbf{Y}$ whatsoever. 
Furthermore, when $\gamma=0$, both \eqref{eq:lag_beta} and \eqref{eq:lag_alpha} lead to trivial solutions.
However, 
recall that our goal is to extract relations between data $\mathbf{X}$ and $\mathbf{Y}$. As with the dual CCA \cite{hardoon2004canonical}, in order to  avoid such trivial solutions, we invoke a Tikhonov regularization term that leads us to our graph dual (gd) CCA formulation  
\begin{subequations}
	\label{eq:gccan}
	\begin{align}
	\max_{\bm\alpha,\,\bm{\beta}}~&~~\bm{\alpha}^\top\!\left(\mathbf{X}^\top\mathbf{X}\mathbf{Y}^\top \mathbf{Y} -\gamma \bm{\alpha}^\top\mathbf{X}^\top\mathbf{X}\mathbf{L}_{\mathcal{G}} \mathbf{Y}^\top \mathbf{Y}\right)\!\bm{\beta}\label{eq:gccanc}\\
	\rm{s.\,to }~&~~\bm{\alpha}^\top \mathbf{X}^\top \mathbf{X} \mathbf{X}^\top \mathbf{X} \bm{\alpha} +\epsilon\bm{\alpha}^\top\mathbf{X}^\top \mathbf{X}\bm{\alpha} = 1\label{eq:gccanc1}\\
	~&~	~\bm{\beta}^\top \mathbf{Y}^\top \mathbf{Y} \mathbf{Y}^\top \mathbf{Y} \bm{\beta}+\epsilon\bm{\beta}^\top\mathbf{Y}^\top \mathbf{Y}\bm{\beta} = 1\label{eq:gccanc2}.
	\end{align}
\end{subequations}
Here, the coefficient $\epsilon>0$ is a pre-selected penalty parameter. Appealing to Lagrange duality theory again, the minimizers $\bm{\alpha}^\ast $ and $ \bm{\beta}^\ast$ are the eigenvectors of
\eqref{eq:gdccas1} and \eqref{eq:gdccas2}  
associated with the largest eigenvalue, namely $(\lambda_1^\ast)^2$; that is, 
	\begin{subequations}
		\label{eq:gdccas}
		\begin{align}
		&(\mathbf{I}-\gamma \mathbf{L}_{\mathcal{G}}) \mathbf{Y}^{\top}\mathbf{Y} (\mathbf{Y}^{\top}\mathbf{Y} + \epsilon \mathbf{I})^{-1} 
		(\mathbf{I}-\gamma \mathbf{L}_{\mathcal{G}}) \mathbf{X}^{\top}\mathbf{X} \bm{\alpha}^\ast \nonumber\\
		&= (\lambda^\ast)^2 	(\mathbf{X}^{\top}\mathbf{X} + \epsilon \mathbf{I}) \bm{\alpha}^\ast	\label{eq:gdccas1}\\
		&(\mathbf{I}-\gamma\mathbf{L}_{\mathcal{G}}) \mathbf{X}^{\top}\mathbf{X} (\mathbf{X}^{\top}\mathbf{X} + \epsilon \mathbf{I})^{-1} (\mathbf{I}-\gamma\mathbf{L}_{\mathcal{G}})\mathbf{Y}^{\top}\mathbf{Y} \bm{\beta}^\ast\nonumber\\
		&= (\lambda^\ast)^2 	(\mathbf{Y}^{
			\top}\mathbf{Y} + \epsilon \mathbf{I})\bm{\beta}^\ast.\label{eq:gdccas2}
		\end{align}
	\end{subequations}
	Moreover, the optimal objective function value coincides with $\lambda_1^\ast$.

When looking for $d$ pairs of dual vectors  $\{(\bm{\alpha}_i,\,\bm{\beta}_i)\}_{i=1}^d$, which are collected to form matrices  $\mathbf{A}:=[\bm{\alpha}_1~\cdots~\bm{\alpha}_d]$ and $\mathbf{B}:=[\bm{\beta}_1~\cdots~\bm{\beta}_d]$, 
our gdCCA becomes
\begin{subequations}\label{eq:gdccam}
	\begin{align}
		\max_{\mathbf{A},\,\mathbf{B}}~&~{\rm Tr}\!\left(\mathbf{A}^\top\mathbf{X}^\top\mathbf{X}\mathbf{Y}^\top \mathbf{Y}\mathbf{B} -\!\gamma \mathbf{A}^\top\mathbf{X}^\top\mathbf{X}\mathbf{L}_{\mathcal{G}} \mathbf{Y}^\top \mathbf{Y}\mathbf{B} \right)\label{eq:gdccamc}\\
		\rm{s.\,to }~&~\mathbf{A}^\top \mathbf{X}^\top \mathbf{X} \mathbf{X}^\top \mathbf{X} \mathbf{A}+\epsilon\mathbf{A}^\top\mathbf{X}^\top \mathbf{X}\mathbf{A} = \mathbf{I}\label{eq:gdccam1}\\
		~&~	\mathbf{B}^\top \mathbf{Y}^\top \mathbf{Y} \mathbf{Y}^\top \mathbf{Y} \mathbf{B}+\epsilon\mathbf{B}^\top\mathbf{Y}^\top \mathbf{Y}\mathbf{B} = \mathbf{I}\label{eq:gdccam2}
	\end{align}
\end{subequations} 
for which the $i$-th column of its optimal solution $\mathbf{A}^\ast$ ($\mathbf{B}^\ast$) is provided by  the generalized eigenvector in \eqref{eq:gdccas1} [\eqref{eq:gdccas2}] associated with the $i$-th largest generalized eigenvalue. Once $\mathbf{A}^\ast$, $\mathbf{B}^\ast$ are found, the optimal canonical vectors sought can be obtained via \eqref{eq:express} as $\mathbf{U}^\ast=\mathbf{X}\mathbf{A}^\ast$ and $\mathbf{V}^\ast=\mathbf{Y}\mathbf{B}^\ast$.


\section{KCCA over Graphs}\label{sec:gkcca}
Although linear models are attractive due to their simplicity,
they cannot capture complex nonlinear data dependencies that are common in real-world applications, including genomics \cite{2003kcca}, functional MRI \cite{blaschko2011semi}, and acoustic feature learning \cite{andrew2013deep}.

Going beyond linearity, we generalize our linear models of CCA over graphs in Sections \ref{sec:gcca} and \ref{sec:gdcca}
to take into account  nonlinear relationships between data $\mathbf{X}$ and $\mathbf{Y}$ 
using kernel methods. In this context, a graph (g)  KCCA framework is developed. We begin with transforming the two datasets using two nonlinear functions to higher (possibly infinite) dimensional feature spaces, and subsequently find low-dimensional canonical variables. Specifically, let $\bm{\phi}_x$ be a mapping from space $\mathbb{R}^{D_x}$ to space $\mathbb{R}^{D_{h}}$ (possibly with $D_{h}=\infty$). It is clear from \eqref{eq:gdccam} that both the objective and the constraints depend on the data $\mathbf{X}$ only through the similarities $\{\langle \mathbf{x}_i,\, \mathbf{x}_j\rangle\}_{i,\,j=1}^N$. Therefore, upon `lifting' all data vectors $\{\mathbf{x}_i\}_{i=1}^N$ to obtain $\{\bm{\phi}(\mathbf{x}_i)\}_{i=1}^N$, all similarities $\{\langle \mathbf{x}_i, \,\mathbf{x}_j\rangle\}_{i,j=1}^N$ can be readily replaced with $\{\langle \bm{\phi}(\mathbf{x}_i),\, \bm{\phi}(\mathbf{x}_j)\rangle\}_{i,j=1}^N$. Nonetheless, evaluating $\{\langle \bm{\phi}(\mathbf{x}_i), \,\bm{\phi}(\mathbf{x}_j)\rangle\}_{i,j=1}^N$ 
can be computationally intractable due to the high-dimensionality. 

To circumvent the cost of explicitly working in the high-dimensional space, the so-called `kernel trick' is employed \cite{RKHS}. 
To this end, we select some kernel function $\kappa_x$, such that $\kappa_x(\mathbf{x}_i, \,\mathbf{x}_j) := \langle\bm{\phi}_x(\mathbf{x}_i),\, \bm{\phi}_x(\mathbf{x}_j)\rangle$ for all $i,\,j=1,\,2,\,\ldots,\,N$, which form the 
$(i,\,j)$-th entries of the so-termed kernel matrix $\bar{\mathbf{K}}_x\in\mathbb{R}^{N\times N}$. Similarly, we can build the kernel matrix $\bar{\mathbf{K}}_y\in\mathbb{R}^{N\times N}$ for data $\mathbf{Y}$ using a different kernel function $\kappa_y$.
As in linear gCCA and gdCCA discussed is Sections \ref{sec:gcca} and \ref{sec:gdcca}, we require that the data in the mapped feature spaces $\{\bm\phi_x(\mathbf{x}_i)\}_{i=1}^N$ and $\{\bm\phi_y(\mathbf{y}_i)\}_{i=1}^N$ be centered, where $\bm\phi_y(\mathbf{y}_i)$ is the nonlinear mapping for `lifting' data $\mathbf{y}_i$ to render kernel matrix $\mathbf{K}_y$. Using the kernel trick again, the required centering in the high-dimensional space can be realized by centering the kernel matrix for data $\mathbf{X}$ as
	\begin{align}
	\label{eq:kxc}
	\mathbf{K}_x(i,j) &:=  \bar{\mathbf{K}}_x(i,j)  - \frac{1}{N} \sum_{\ell=1}^N \bar{\mathbf{K}}_x(\ell, j)-\frac{1}{N}\sum_{\ell=1}^N \bar{\mathbf{K}}_x(i,\ell) \nonumber\\
	&\quad ~+\frac{1}{N^2}\sum_{m=1}^N\sum_{n=1}^N \bar{\mathbf{K}}_x(m,n)
	\end{align}
	and likewise for centering $\mathbf{K}_y$.

Upon replacing $\mathbf{X}^{\top}\mathbf{X}$ and $\mathbf{Y}^{\top}\mathbf{Y}$ in \eqref{eq:gdccam}  with $\mathbf{K}_x$ and $\mathbf{K}_y$, we arrive at our gKCCA 
\begin{subequations}
	\label{eq:gkccam}
	\begin{align}
	\max_{\mathbf{A},\, \mathbf{B}}~&~~ {\rm{Tr}}(\mathbf{A}^{\top}\mathbf{K}_x\mathbf{K}_y \mathbf{B}-\gamma \mathbf{A}^{\top}\mathbf{K}_x \mathbf{L}_{\mathcal{G}}\mathbf{K}_y\mathbf{B})\label{eq:gkccamc}\\
	\rm{s. \,to }~&~~\mathbf{A}^{\top}\mathbf{K}_x^2\mathbf{A} + \epsilon \mathbf{A}^{\top} \mathbf{K}_x \mathbf{A}=\mathbf{I}\label{eq:gkccamc1}\\
	~&~~\,\mathbf{B}^{\top}\mathbf{K}_y^2\mathbf{B} + \epsilon \mathbf{A}^{\top} \mathbf{K}_y \mathbf{B}=\mathbf{I}\label{eq:gkccamc2}.
	\end{align}
\end{subequations}

It is clear that with properly selected kernel matrices $\mathbf{K}_x$ and $\mathbf{K}_y$, gKCCA is able to capture nonlinear correlations between $\mathbf{X}$ and $\mathbf{Y}$, while also leveraging the graph prior information of the common sources. Following the steps used  to solve the gCCA problem \eqref{eq:gpcca}, the solution to \eqref{eq:gkccam} is summarized in Theorem \ref{th:gkcca}, with its proof deferred to Appendix \ref{profgkcca}. The main steps of the gKCCA are listed in Alg. \ref{alg:gkcca}. 
\begin{theorem}
	\label{th:gkcca}
	If $\mathbf{K}_x$ and $\mathbf{K}_y$ are nonsingular, the optimal solutions $\mathbf{A}^{\ast}$ and $\mathbf{B}^{\ast}$ to \eqref{eq:gkccam} are given by
	\begin{subequations}
		\begin{align}
		\label{eq:astar2}
		\mathbf{A}^{\ast}&:=\mathbf{K}_x^{-1/2}(\mathbf{K}_x+\epsilon \mathbf{I})^{-1/2}\bar{\mathbf{A}}^{\ast}\\
		\label{eq:bstar2}
		\mathbf{B}^{\ast}&:=\mathbf{K}_y^{-1/2}(\mathbf{K}_y+\epsilon \mathbf{I})^{-1/2}\bar{\mathbf{B}}^{\ast}
		\end{align}
	where matrices $\bar{\mathbf{A}}^{\ast}\in\mathbb{R}^{N\times d}$ and $\bar{\mathbf{B}}^{\ast}\in\mathbb{R}^{N\times d}$ collect as columns the top $d$ left and right singular vectors of 
	\begin{align}
	\label{eq:gkccack}
 \mathbf{C}:=(\mathbf{K}_x+\epsilon\mathbf{I})^{-1/2}\mathbf{K}_x^{1/2}(\mathbf{I}-\gamma \mathbf{L}_{\mathcal{G}})\mathbf{K}_y^{1/2}(\mathbf{K}_y+\epsilon\mathbf{I})^{-1/2}.
	\end{align}
		\end{subequations}
	 Furthermore, the optimal objective value \eqref{eq:gkccamc} is the sum of the $d$ largest singular values of 
	  $\mathbf{C}$.
\end{theorem}

\begin{algorithm}[t]
	\caption{Graph kernel canonical correlation analysis}
	\label{alg:gkcca}
	\begin{algorithmic}[1]
		\STATE {\bfseries Input:}		
$\{\mathbf{x}_i\}_{i=1}^N$, $\{\mathbf{y}_i\}_{i=1}^N$, $\mathbf{W}$, $d$, $\gamma$, $\epsilon$, $\kappa_x(\cdot)$, and $\kappa_y(\cdot)$.
		\STATE {\bfseries Construct} $\mathbf{K}_x$ and $\mathbf
		{K}_y$
		 using \eqref{eq:kxc}.
		\STATE {\bfseries Build} $\mathbf{L}_{\mathcal{G}}$ using \eqref{eq:lg}.
		\STATE {\bfseries Perform} SVD on $\mathbf{C}:=\mathbf{U}\bm{\Sigma}\mathbf{V}^{\top}$ in \eqref{eq:gkccack},
		where the diagonal elements of $\bm{\Sigma}$ are organized in descending order; ${\mathbf{U}}\in\mathbb{R}^{N\times N}$, ${\mathbf{V}} \in \mathbb{R}^{N \times N}$, and $\bm{\Sigma}\in \mathbb{R}^{N \times N}$.
		\STATE{\bfseries Extract} the first $d$ columns of ${\mathbf{U}}$ and ${\mathbf{V}}$ to form $\bar{\mathbf{A}}^\ast\in\mathbb{R}^{N\times d}$ and $\bar{\mathbf{B}}^\ast\in\mathbb{R}^{N\times d}$, respectively.
		\STATE{\bfseries Compute} $\mathbf{A}^{\ast}=\mathbf{K}_x^{-1/2}(\mathbf{K}_x+\epsilon \mathbf{I})^{-1/2}\bar{\mathbf{A}}^\ast$ and $\mathbf{B}^{\ast}=\mathbf{K}_y^{-1/2}(\mathbf{K}_y+\epsilon \mathbf{I})^{-1/2}\bar{\mathbf{B}}^\ast$.
		\STATE {\bfseries Output:}\label{step:5} $\mathbf{A}^{\ast}$ and $\mathbf{B}^{\ast}$.
	\end{algorithmic}
\end{algorithm}

\begin{remark}
	When the kernel functions needed to form $\mathbf{K}_x$ and $\mathbf{K}_y$ are not available, one may presume $\mathbf{K}_{x}:=\sum_{m=1}^M\theta_m \mathbf{K}_{m}$ and $\mathbf{K}_{y}:=\sum_{m=1}^M \delta_m\mathbf{K}_{m}$ for \eqref{eq:gkccam}. Here, $\{\mathbf{K}_m\}_{m=1}^M$ are known kernel matrices for a preselected dictionary of kernels, while $\{\theta_m,\,\delta_m\}_{m=1}^M$ are unknown coefficients to be optimized along with the canonical vectors through \eqref{eq:gkccam}. Such a data-driven approach is also known as  multi-kernel learning, and it has been broadly studied; see for example, \cite{mkl2004, meng2017}.
\end{remark}

In terms of computational cost, we summarize the complexities of gCCA, gdCCA, gKCCA, CCA, dCCA, and KCCA in Table \ref{tab:comp}, where $D:=\max(D_x,D_y)$. Note  that gCCA incurs higher computational cost than standard CCA, due to the extra multiplication term of $\mathbf{Y}\mathbf{L}_{\mathcal{G}}\mathbf{X}^T$ in gCCA.
If  $N\ll D$, then gCCA in its present form is not feasible, or suboptimal even if the pseudo-inverse or Tikhonov regularization is employed, at  computational complexity  $\mathcal{O}(D^3)$. In this case, gdCCA is computationally more attractive since its complexity grows only linearly with $D$. In terms of gKCCA,
when $D\gg N$, evaluating the kernel matrices dominates the computational complexity, giving rise to $\mathcal{O}(DN^2)$. When $D\ll N$, Steps $4$ and $6$ in Alg. \ref{alg:gkcca} dominate the complexity, incurring complexity of $\mathcal{O}(N^3)$.

\begin{table}[]
	\renewcommand\arraystretch{1.3}
	\centering
	\caption{Computational complexity comparison}
	\label{tab:comp}
	\begin{tabular}{|c|c|}
		\hline
		gCCA  
		& $\mathcal{O}(\min(D_x,D_y)N^2)$\\
		\hline
		CCA   
		& $\mathcal{O}(D^2N)$  \\
		\hline
		gdCCA (dCCA)  
		& $\mathcal{O}(D N^2)$\\
		\hline
		gKCCA (KCCA)   
		& $\mathcal{O}(\max(D, N)N^2)$ \\
		\hline
	\end{tabular}
\end{table}

\section{Numerical Tests}\label{sec:test}
To showcase the merits of our novel approaches, several classification experiments using real data are reported in this section. Classification accuracies of our proposed gCCA, gdCCA and gKCCA are compared with competing alternatives.

\subsection{Tests for gCCA}
\label{sec:e1}

In this experiment, the AR face dataset \cite{ARface}, and the Extended Yale-B (EYB) face image dataset \cite{yaleb}, were used to examine the classification performance of different schemes, including gCCA, CCA, graph (g)  PCA \cite{gpca}, PCA, graph regularized multi-set (GrM) CCA \cite{pr2014sun}, and the $k$-nearest neighbors (KNN) method.

The AR face database contains color face images of $100$ individuals, each depicted in  $26$ images. These  $26$ images per person were taken under different lighting conditions, occlusions and expressions. Each image was cropped and resized to $40 \times 30$ pixels, converted to grayscale image, and vectorized to obtain a $1,200\times 1$ vector. The $1,200$ features of each image are unevenly split into two views, where one view consists of the first $300$ features collected in one column of $\mathbf{X}_0\in\mathbb{R}^{300\times 2,600}$ ($2,600$ columns for all the images) , while the remaining  $900$ features were used to form  $\mathbf{Y}_0\in\mathbb{R}^{900\times 2,600}$. Suppose that $N_{\rm tr}$ columns  were randomly drawn from $26$ columns of $\mathbf{X}_0$ and $\mathbf{Y}_0$ that correspond to one person, to form the training data $\mathbf{X}\in\mathbb{R}^{300\times 100N_{\rm tr}}$ and $\mathbf{Y}\in\mathbb{R}^{900\times 100N_{\rm tr}}$, respectively. For the remaining $(26-N_{\rm tr})$ columns of $\mathbf{X}_0$ associated with each person, half of them will be used for tuning the hyper-parameters, and the other  half for testing, which are collected in $\mathbf{X}_{\rm tu}\in\mathbb{R}^{300\times 100(13-0.5N_{\rm tr})}$ and $\mathbf{X}_{\rm te}\in\mathbb{R}^{300\times 100(13-0.5N_{\rm tr})}$ accordingly. Here, we consider the scenario where only one view, namely $\mathbf{X}_{\rm te}$, is available in the testing phase, which is of practical importance when one only has partial information about the testing images.

The EYB database consists of frontal face images of $38$ individuals, each of which has around $65$ color images of $192\times 168$ pixels. All images are resized to $30\times 20$ pixels and converted to grayscale before being  vectorized to obtain a $600\times 1$ vector. Then, the vector associated with each image is split into two subvectors (views) with $D_x=250$ and $D_y=350$. For each individual, $N_{\rm tr}$ images are randomly selected and the corresponding two views are used to construct the training datasets $\mathbf{X}\in\mathbb{R}^{D_x \times 38N_{\rm tr}}$ and $\mathbf{Y}\in\mathbb{R}^{D_y \times 38N_{\rm tr}}$. Among the remaining images, $(30-0.5N_{\rm tr})$ images per individual are used for tuning dataset $\mathbf{X}_{\rm tu}\in\mathbb{R}^{D_x \times 38(30-0.5N_{\rm tr})}$ and another $(30-0.5N_{\rm tr})$ for testing dataset $\mathbf{X}_{\rm te}\in\mathbb{R}^{D_x \times 38(30-0.5N_{\rm tr})}$, after following a similar process to build $\mathbf{X}$.

Letting $N:=100N_{\rm tr}$ for the AR data experiment ($N:=38N_{\rm tr}$ for EYB), we collected all common sources $\{\mathbf{s}_i \}_{i=1}^N$ into matrix $\mathbf{S}$, which was constructed using the training data as follows:  $\mathbf{S}:=[\mathbf{X}^{\top}\,\mathbf{Y}^{\top}]^{\top}=[\mathbf{s}_1\,\cdots\, \mathbf{s}_N]$. Based on $\mathbf{S}$, matrix  $\mathbf{W}$ is formed to have  $(i,\,j)$-th entry given by
\begin{equation}\label{eq:Wi}
w_{ij}:=
\begin{cases}
\frac{\mathbf{s}_i^{\top}\mathbf{s}_j}{\|\mathbf{s}_i\|_2\|\mathbf{s}_j\|_2} &\mbox{$\mathbf{s}_i\in{\mathcal{N}}_{k}(\mathbf{s}_j) {\rm{~or ~}}\mathbf{s}_j\in{\mathcal{N}}_{k}(\mathbf{s}_i)$}\\
0 &\mbox{otherwise}
\end{cases}
\end{equation}
for $i,\,j=1,\,2,\,\ldots,\, N$, where $\mathcal{N}_k(\mathbf{s}_j)$ denotes the set of the $k$-nearest neighbors of $\mathbf{s}_j$ that belong to the same class (person) in $\mathbf{S}$. In this experiment, $k=N_{\rm tr}-1$ was kept fixed.

In this experiment, $30$ Monte Carlo (MC) simulations were run to assess the  classification performance of gCCA, standard CCA, GrMCCA, gPCA, PCA, and KNN on  the AR face dataset, as well as the EYB dataset. For fairness, the weight matrix $\mathbf{W}$ in \eqref{eq:Wi} is used for gPCA. The classification accuracy is defined as the ratio between the number of correctly classified images and the total number of images tested. For gCCA, CCA, GrMCCA, gPCA, and PCA, $50$ $(100)$ canonical vectors for the AR (EYB) face dataset were found to obtain the low-dimensional representations of testing data, which were subsequently classified through the $10$-nearest neighbors algorithm based on the Euclidean distance metric. The hyper-parameters in gCCA, gPCA, and GrMCCA were tuned among $30$ logarithmically-spaced values between $10^{-3}$ and $10^3$ to maximize the classification accuracies on `tuning set' of images. 

\begin{figure}[t]
	\centering 
	\includegraphics[scale=0.6]{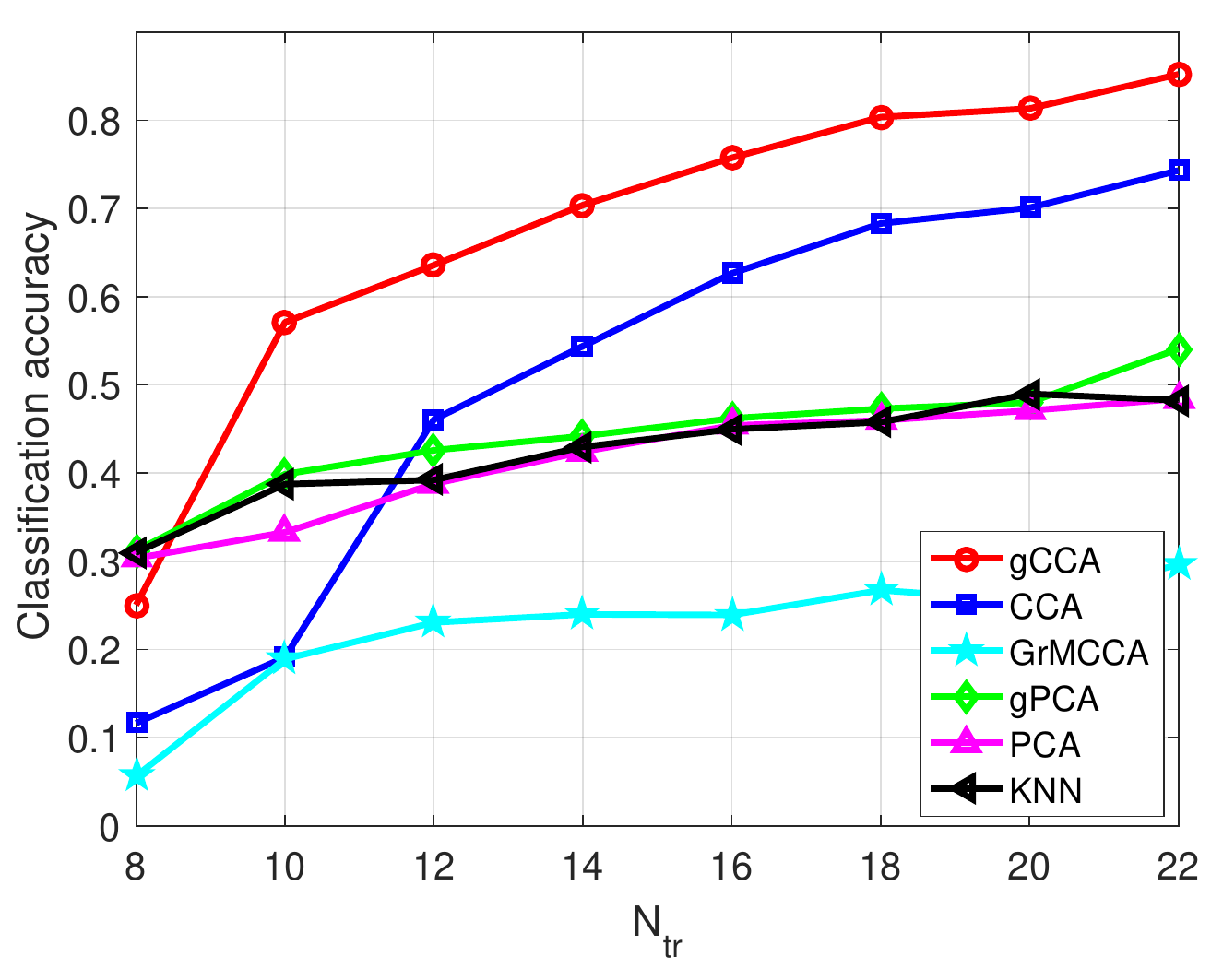} 
	\caption{\small{Classification accuracy of gCCA on the AR face dataset \cite{ARface}.}}
	\label{fig:gcca_fig1}
\end{figure}

\begin{figure}[t]
	\centering 
	\includegraphics[scale=0.6]{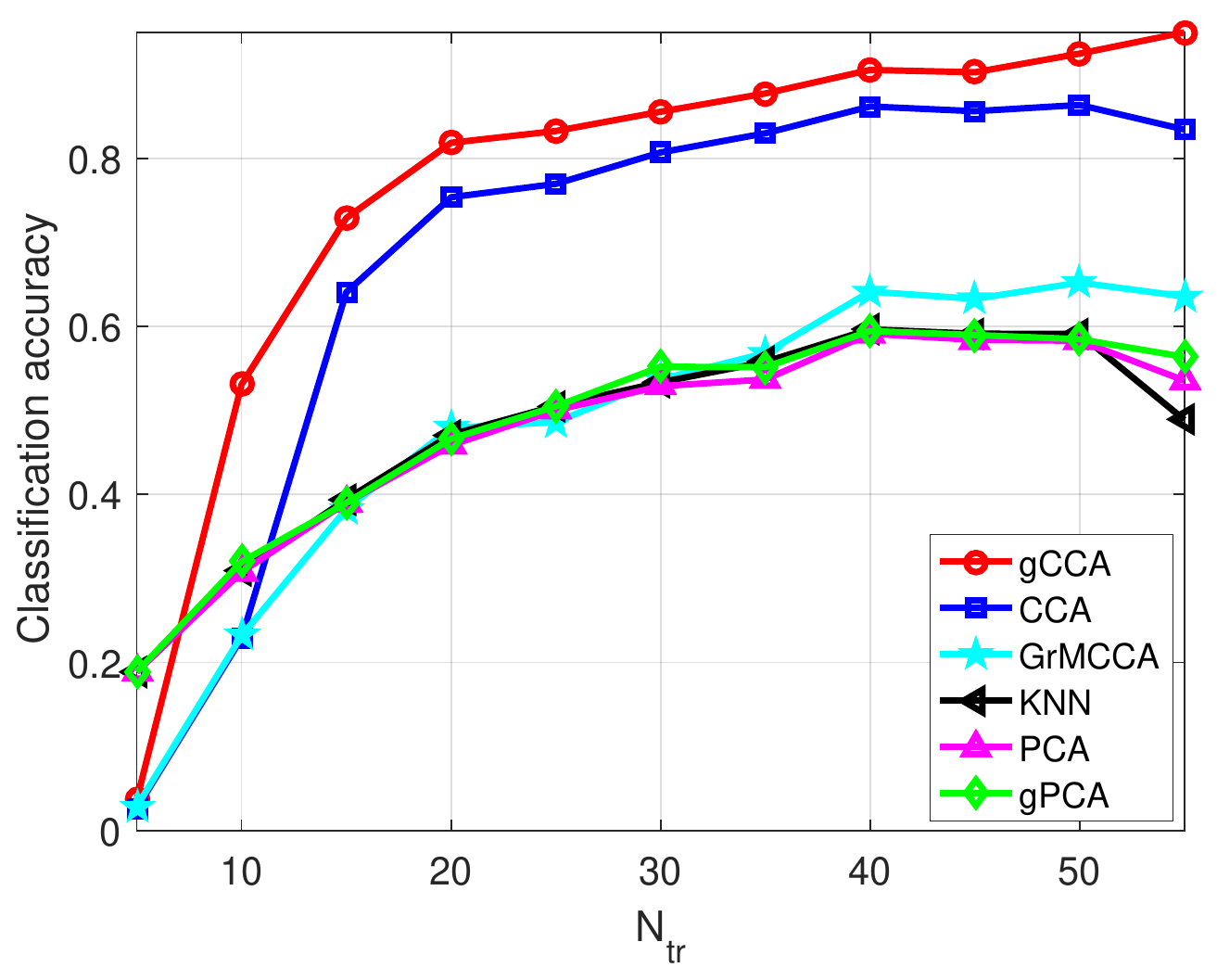} 
	\caption{\small{Classification accuracy of gCCA on the EYB dataset \cite{yaleb}.}}
	\label{fig:gcca_fig2}
\end{figure}

Figures \ref{fig:gcca_fig1} and \ref{fig:gcca_fig2} depict the classification accuracies of gCCA, CCA, GrMCCA, gPCA, PCA, and KNN on the AR data, and the EYB data, respectively, for a varying number of training samples. It is evident that the accuracies of all simulated schemes improve as $N_{\rm tr}$ grows, and our proposed gCCA outperforms  alternatives for  $N_{\rm tr}\ge 10$. This corroborates that incorporating the source graph that encodes dependencies among common sources, pays off.


\subsection{Tests for gdCCA}
The second experiment evaluates the capability of gdCCA for classification using again  the AR face dataset and the EYB dataset.
Per MC run on the AR face dataset, we collected all images of $10$ randomly sampled people. For each selected person, $N_{\rm tr}$, $(13-0.5 N_{\rm tr})$, and $(13-0.5N_{\rm tr})$ images were randomly drawn for training, tunning, and testing, respectively. In the training phase, each image was first converted to a grayscale image, resized to $80\times 60$ pixels, and subsequently lexicographically ordered  to obtain  a $4,800\times 1$ vector. To create the two views, this vector was partitioned into two subvectors of size $D_x=1,000$ for $\mathbf{X}\in\mathbb{R}^{D_x \times 10N_{\rm tr}}$ and of size $D_y=3,800$ for $\mathbf{Y}\in\mathbb{R}^{D_y \times 10N_{\rm tr}}$. Similarly, the training data $\mathbf{X}_{\rm tu}\in\mathbb{R}^{D_x \times 10(13-0.5N_{\rm tr})}$ and testing data $\mathbf{X}_{\rm te}\in\mathbb{R}^{D_x \times 10(13-0.5N_{\rm tr})}$ were generated. 

Per realization on the EYB dataset, images of $10$ individuals were randomly selected, and the two-view data $\mathbf{X}\in\mathbb{R}^{D_x \times 10N_{\rm tr}}$ and $\mathbf{Y}\in\mathbb{R}^{D_y\times 10N_{\rm tr}}$ were generated using the same procedure described for the AR data, except for
 $D_x=1,000$ and $D_y=7,000$. For both the tuning data $\mathbf{X}_{\rm tu}\in\mathbb{R}^{D_x\times 10(30-0.5N_{\rm tr})}$ and the testing data $\mathbf{X}_{\rm te}\in\mathbb{R}^{D_x\times 10(30-0.5N_{\rm tr})}$, a number of $(30-0.5N_{\rm tr})$ images were randomly chosen per person.

The two-view data in the training phase form $\mathbf{S}=[\mathbf{X}^{\top}\;\mathbf{Y}^{\top}]^{\top}$ and are further used to build $\mathbf{W}$ as in \eqref{eq:Wi}. For fairness, graph dual (gd)  PCA \cite{gpca} is tested with the same $\mathbf{W}$ as in gdCCA. Moreover, the two associated graph adjacency matrices in Laplacian regularized (Lr) CCA \cite{blaschko2011semi} are constructed via \eqref{eq:Wi} after substituting $\mathbf{S}$ by $\mathbf{X}$ and $\mathbf{Y}$, respectively. We tune the hyper-parameters in gdCCA, dual (d) CCA, LrCCA and  gdPCA among $30$ logarithmically spaced values between $10^{-3}$ and $10^3$ to maximize the classification accuracy on data $\mathbf{X}_{\rm tu}$. Here, dCCA is implemented by gdCCA after assigning $\gamma=0$. In gdCCA, dCCA, LrCCA, gdPCA and dPCA \cite{gpca}, 20 and 100 projection vectors are used for obtaining lower-dimensional representations of $\mathbf{X}_{\rm te}$ for AR data and EYB data, respectively. Then, the K-NN rule with $K=10$ was  applied to carry out the classification tasks.

\begin{figure}[t]
	\centering 
	\includegraphics[scale=0.6]{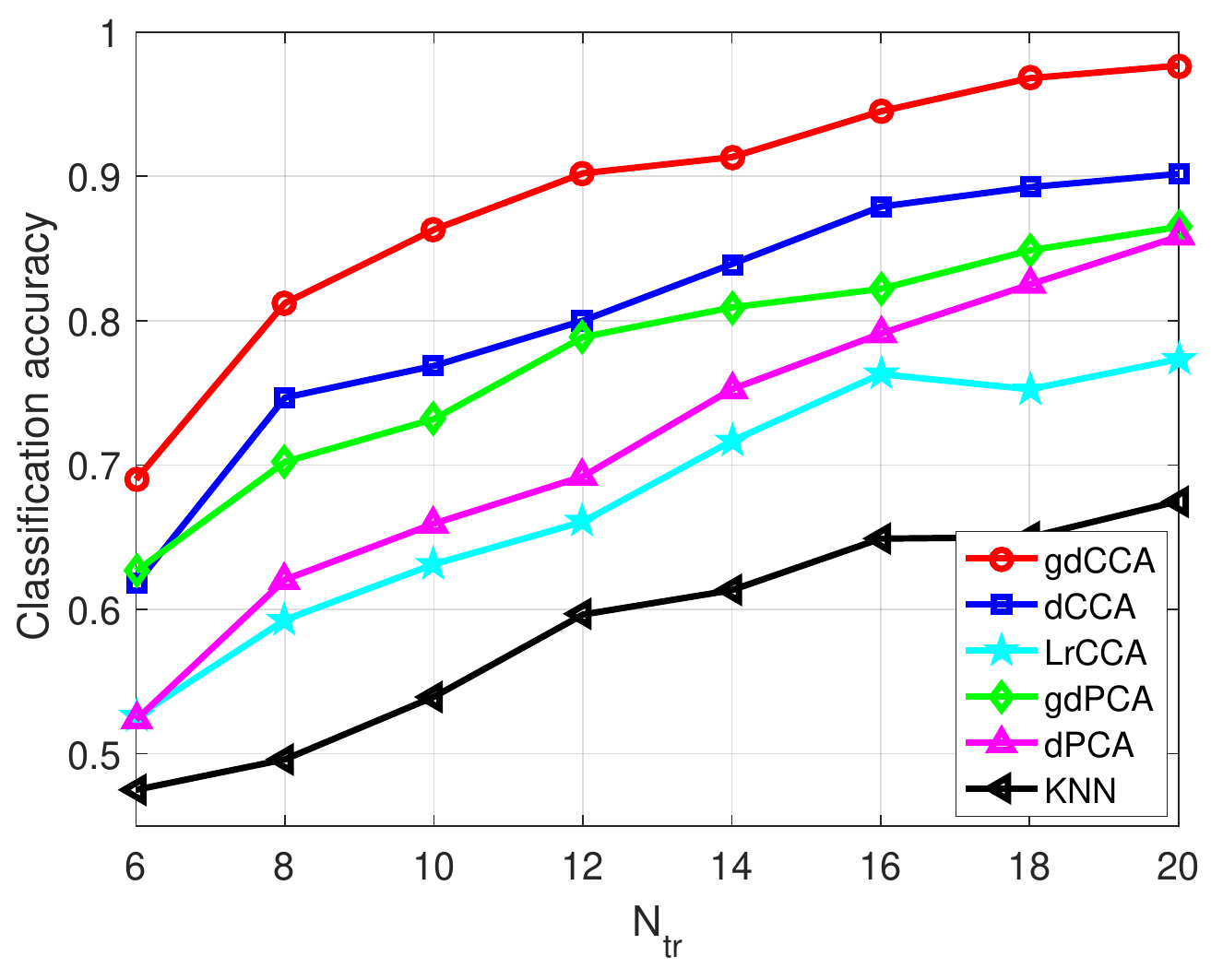} 
	\caption{\small{Classification accuracy of gdCCA using dataset \cite{ARface}.}}
	\label{fig:gdcca_fig1}
\end{figure}

\begin{figure}[t]
	\centering 
	\includegraphics[scale=0.6]{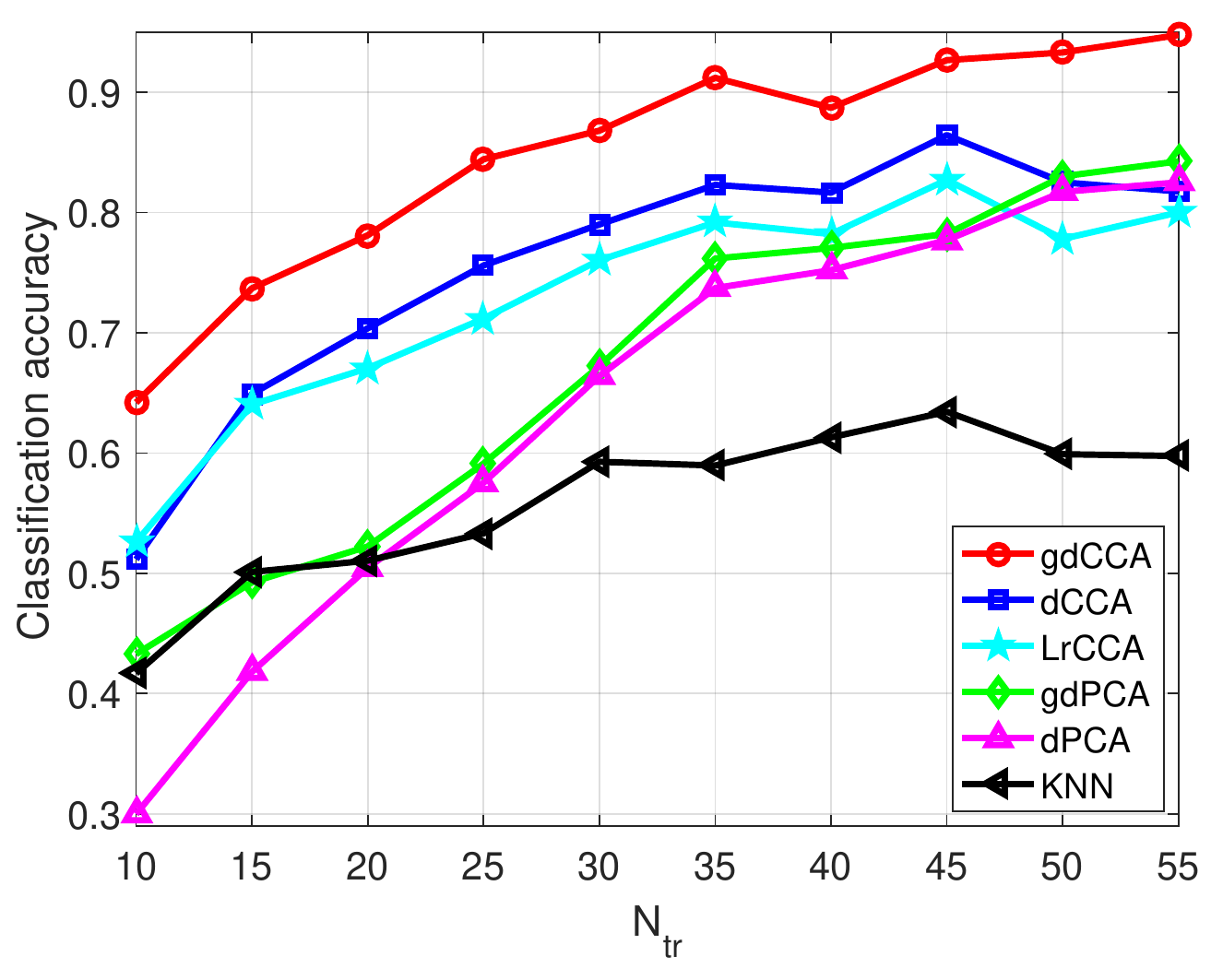} 
	\caption{\small{Classification accuracy of gdCCA using dataset \cite{yaleb}.}}
	\label{fig:gdcca_fig2}
\end{figure}

Figures \ref{fig:gdcca_fig1} and \ref{fig:gdcca_fig2} present the averaged classification accuracies of gdCCA, dCCA, LrCCA, gdPCA, dPCA, and KNN for a varying number of training images per person over $30$ MC realizations. Clearly,  
our gdCCA enjoys the best classification performance among all simulated schemes for different training samples.

\begin{figure}[t]
	\centering 
	\includegraphics[scale=0.6]{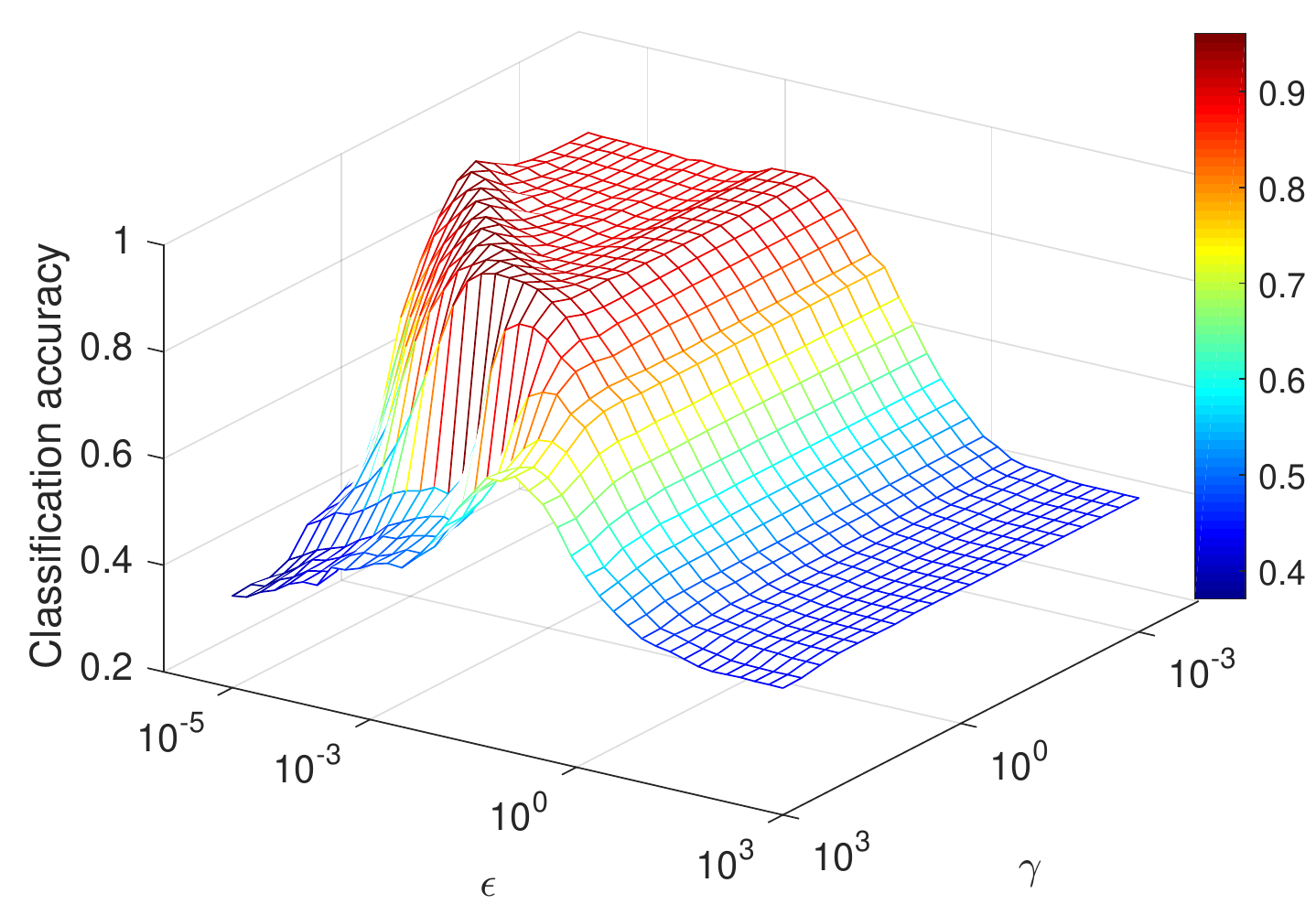} 
	\caption{\small{Classification accuracy of gdCCA versus $\gamma$ and $\epsilon$.}}
	\label{fig:gdcca_fig3}
\end{figure}

There are two hyper-parameters, namely $\gamma$ and $\epsilon$ in gdCCA. To  understand how the hyper-parameters influence the classification performance, the gdCCA was simulated on the AR face dataset for a range of $\gamma$ and $\epsilon$ values. 
For each person, $17$ ($9$) images were employed for training (testing). 
Figure \ref{fig:gdcca_fig3} plots the averaged classification accuracies over $30$ MC runs, with $\gamma$ varying from $10^{-3}$ to $10^3$ and $\epsilon$ from $10^{-5}$ to $10^3$. For  small $\gamma$ values, the performance of gdCCA with small $\epsilon$ values outperforms that using large $\epsilon$ values. This is because  with small $\gamma$, gdCCA approximates dCCA, and the Tikhonov regularization with excessively large $\epsilon$ values dominates the term for promoting uncorrelatedness between canonical variables.
When $\epsilon$ is small, with $\gamma$ increasing, the classification accuracy gradually increases by progressively exploiting the graph information, but subsequently decreases due to discarding the maximization of  canonical correlations.  
 Those observations confirm the assertion that with properly selected and nonzero $\gamma$ and $\epsilon$ values, the performance of gdCCA reaches the best, in which case both maximizing the canonical correlations and exploiting the graph knowledge are in effect.


\subsection{Tests for gKCCA}\label{sec:e3}
This last experiment assesses gKCCA for classification using the MNIST dataset \footnote{Downloaded from \url{http://yann.lecun.com/exdb/mnist/}.}. 
There are $10$ classes of hand-written $28\times 28$ grayscale digit images in the MNIST, and each class (digit) consists of $7,000$ images. Per MC run, $5$ classes of images were randomly sampled for classification. For each selected class, $N_{\rm tr}$,  $0.5N_{\rm tr}$, and $0.5N_{\rm tr}$ images are randomly sampled for training, parameter tuning, and testing, respectively. The two-view data were created as follows.
The images were first resized to $20\times 20$ pixels, followed by vectorization. Each vector was split to $2$ subvectors of sizes $D_x$ and $D_y=400-D_x$ for the two views. The first/second view of training data is denoted by training dataset $\mathbf{X}\in\mathbb{R}^{D_x\times 5N_{\rm tr}}/\mathbf{Y}\in\mathbb{R}^{D_y\times 5N_{\rm tr}}$. The tuning/testing dataset $\mathbf{X}_{\rm tu}/\mathbf{X}_{\rm te}$ are the first views of tuning/testing images.

Gaussian kernels were used for $\mathbf{X}$, $\mathbf{Y}$, and the common source $\mathbf{S}:=[\mathbf{X}^{\top}\;\mathbf{Y}^{\top}]^{\top}$, 
 whose bandwidth parameters were set as the medians of the corresponding Euclidean distances. 
  The idea to generate the $\mathbf{W}$ in Sec. \ref{sec:e1} was adopted and adjusted for constructing the graph adjacency matrix, which was also denoted by $\mathbf{W}$ for notational simplicity. Obviously, the similarity between two sources in $\mathbf{S}$ can not be measured by the linear correlation coefficient, which instead can be represented by a corresponding element in the kernel matrix of $\mathbf{S}$, namely $\mathbf{K}_s$. Specifically, 
\begin{equation}\label{eq:wik}
w_{ij}:=
\begin{cases}
\mathbf{K}_s(i,j) &\mbox{$\mathbf{s}_i\in{\mathcal{M}}_{k_1}(\mathbf{s}_j)$ or $\mathbf{s}_j\in{\mathcal{M}}_{k_1}(\mathbf{s}_i)$}\\
0 &\mbox{otherwise}
\end{cases}
\end{equation}
for $i,\,j = 1,2,\ldots,5N_{\rm tr}$, where $\mathbf{s}_i$ denotes the $i$-th source ($i$-th column) in $\mathbf{S}$, and ${{\mathcal{M}}}_{k_1}(\mathbf{s}_j)$ is the set containing the $k_1$-nearest neighbors of $\mathbf{s}_j$ from the same class. 
In the simulations of this subsection, $k_1= N_{tr}-1$. Further, graph kernel (gK) PCA \cite{gpca} was simulated with the same $\mathbf{W}$ as in gKCCA. The graph Laplacian regularized (Lr) KCCA \cite{blaschko2011semi} was associated with two graph adjacency matrices, which were obtained by \eqref{eq:wik} after  substituting $\mathbf{K}_s$ with  $\mathbf{K}_x$ and $\mathbf{K}_y$ accordingly. For fairness, all the kernel-based methods, namely  gKCCA, KCCA, LrKCCA, gKPCA, and KPCA, shared the same kernel $\mathbf{K}_x$ (and $\mathbf{K}_y$). When implementing the CCA-based and PCA-based subspace methods, $20$ projection vectors were used for classification using the K-NN  algorithm with $K=10$.  The hyper-parameters of gKCCA, KCCA, gdCCA, dCCA, LrKCCA, LrCCA, gKPCA, and gdPCA, were selected from $30$ logarithmically spaced values between $10^{-3}$ and $10^3$. For each algorithm, the parameters were selected with the best classification accuracy on the tuning dataset $\mathbf{X}_{\rm tu}$. In the following tests, the classification performance of all aforementioned algorithms was achieved after running $30$ independent realizations.

\begin{figure}[t]
	\centering 
	\includegraphics[scale=0.6]{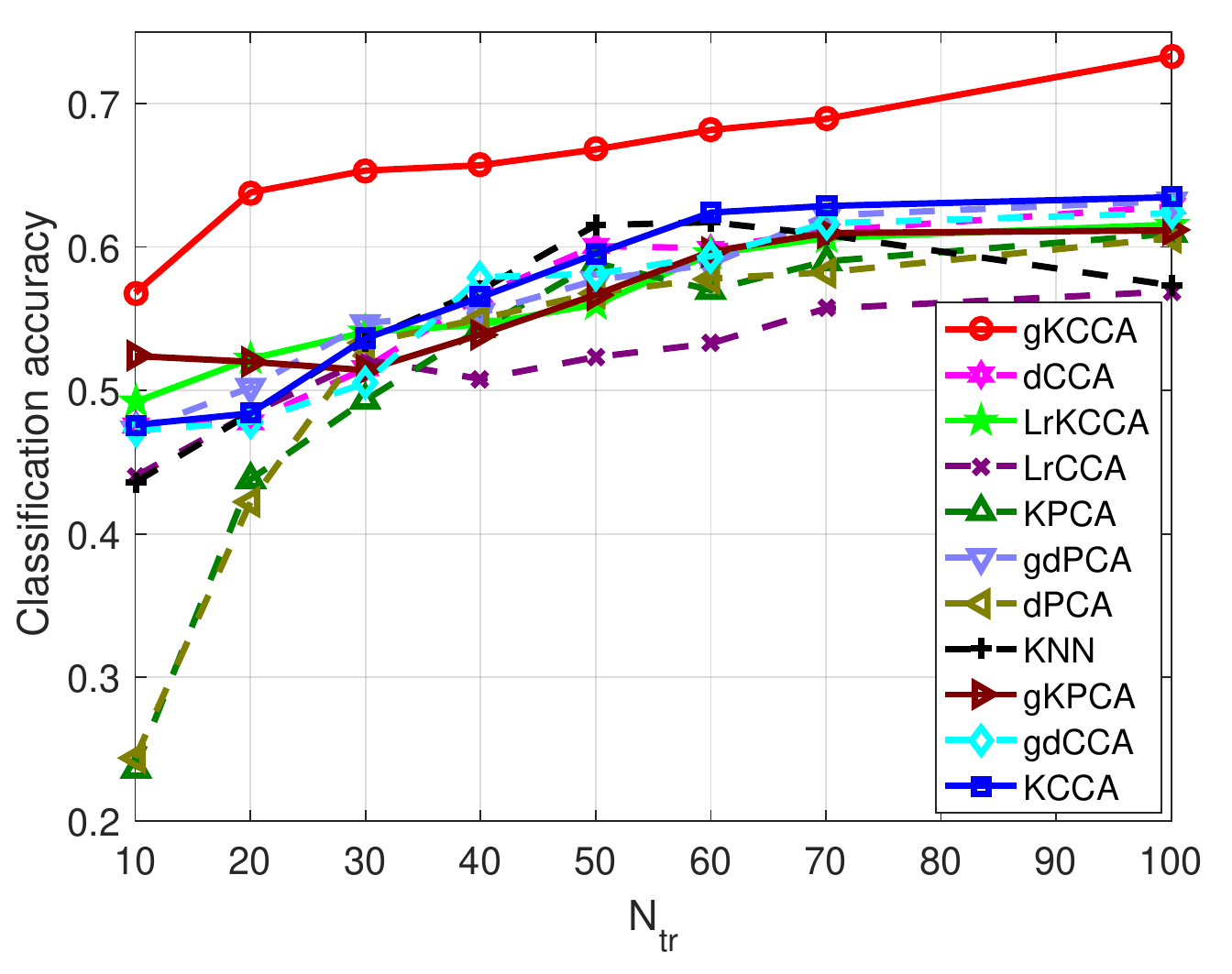} 
	\caption{\small{Classification accuracy of gKCCA versus $N_{\rm tr}$.}}
	\label{fig:gkcca_fig1}
\end{figure}

\begin{figure}[t]
	\centering 
	\includegraphics[scale=0.6]{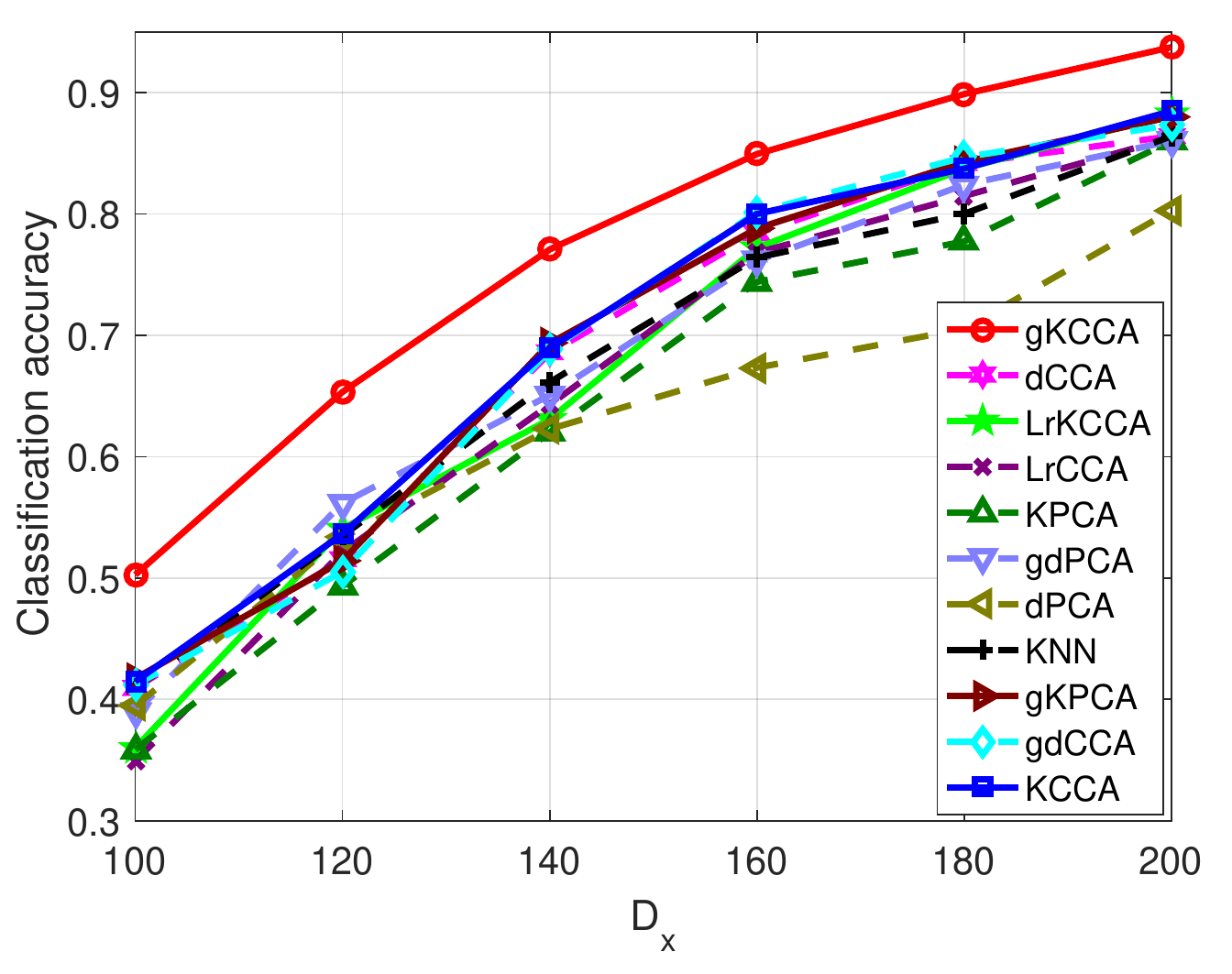} 
	\caption{\small{Classification accuracy of gKCCA versus $D_x$.}}
	\label{fig:gkcca_fig2}
\end{figure}

In Fig. \ref{fig:gkcca_fig1}, the classification accuracies of simulated schemes for a variable  number of training samples are reported, with  $D_x=120$ and $D_y=280$. The plots validate the advantage of our gKCCA relative to the other $10$ methods. 
Moreover, with extra training samples becoming available, the performance of all simulated schemes improves. Figure \ref{fig:gkcca_fig2} depicts the classification accuracies of all methods for different $D_x$ values, with $N_{\rm tr}=30$ kept  fixed. It is clear that gKCCA outperforms alternatives under different vector splittings. On the other hand, with $D_x$ decreasing, it becomes more challenging to classify the testing data, so the classification accuracies of all schemes decrease. Interestingly, the performance gap between gKCCA and the others widens for smaller $D_x$ values.


\section{Conclusions}\label{sec:concl} 

Graph regularized CCA, dual CCA, as well as kernel CCA methods were revisited in this paper to exploit hidden low-dimensional common structures from two-view data of the same sources. Distinguishing itself from prior CCA contributions, our gCCA framework leverages additional information to improve the low-dimensional approximations through the canonical variables, by embedding the hidden common sources in a graph and invoking this graph prior knowledge as a CCA regularizer.  
As such, canonical pairs that are able to capture the structural information between data vectors can be revealed. In certain practical setups where the number of data samples is small relative to the data vector dimensionality, our gCCA is not directly applicable, or leads to suboptimal performance and incurs high computational complexity. To bypass this, the dual model of gCCA, namely gdCCA, is put forth. To further account for nonlinear data dependencies,  the graph kernel CCA is  developed. Numerical tests on several real-world datasets are presented to demonstrate the merits of the novel  approaches.

This paper opens up several intriguing directions for future research. To start, developing data-driven approaches to select the appropriate kernels (graphs) from a given or constructed dictionary of kernels (graphs) is timely and pertinent. To endow the proposed gCCA algorithms with scalability, 
distributed and online implementations are well-motivated for handling large-scale and/or high-dimensional streaming data. Generalizing our gCCA models to unpaired or multi-view datasets constitutes another interesting direction.

\appendix

\subsection{Proof of Theorem~\ref{thm:gcca}}\label{prof2}
Letting
\begin{equation*}
\bar{\mathbf{U}}:=\bm{\Sigma}_x^{1/2}\mathbf{U}\in\mathbb{R}^{D_x\times d},\quad {\rm and}\quad \bar{\mathbf{V}}:={\bm{\Sigma}}_y^{1/2}\mathbf{V}\in\mathbb{R}^{D_y \times d}
\end{equation*}  
the objective function \eqref{eq:gpccac} can be rewritten as
\begin{equation}
\label{eq:gkccamc_2}
{\rm Tr}(\bar{\mathbf{U}}^{\top}\mathbf{C}\bar{\mathbf{V}}):=\textrm{Tr}(\bar{\mathbf{U}}^{\top}\boldsymbol{\Sigma}_x^{-1/2}(\boldsymbol{\Sigma}_{xy}-\gamma\mathbf{X}\mathbf{L}_{\mathcal{G}}\mathbf{Y}^{\top})\boldsymbol{\Sigma}_y^{-1/2}\bar{\mathbf{V}})\nonumber
\end{equation}  
and problem \eqref{eq:gpcca} boils down to
\begin{subequations}
	\label{eq:gkccam_3}
	\begin{align} 
	\max_{\bar{\mathbf{U}}, \, \bar{\mathbf{V}}} ~&~~	{\rm{Tr}}(\bar{\mathbf{U}}^{\top}\mathbf{C}\bar{\mathbf{V}}) \\
	{\rm s.\,to }~&~~ \bar{\mathbf{U}}^{\top}\bar{\mathbf{U}}=\mathbf{I},\quad{\rm and}\quad\bar{\mathbf{V}}^{\top}\bar{\mathbf{V}}=\mathbf{I}. \label{eq:gkccam_3c2}
	\end{align} 
\end{subequations}
Let $\bar{\mathbf{u}}_i \in \mathbb{R}^{D_x\times 1}$ and $\bar{\mathbf{v}}_i\in \mathbb{R}^{D_y \times 1}$ denote the $i$-th column of $\bar{\mathbf{U}}$ and $\bar{\mathbf{V}}$, respectively, with $i=1,\,2,\,\ldots, \,d$.  
The problem in \eqref{eq:gkccam_3} can be solved using $d$ iterations with  each iteration targeting the optimum over $\bar{\mathbf{u}}_i$ and $\bar{\mathbf{v}}_i$, namely 
\begin{subequations}\label{eq:pm}
	\begin{align}
	(\bar{\mathbf{u}}_i^{\ast}, \bar{\mathbf{v}}_i^{\ast}): = \arg\max_{\bar{\mathbf{u}}_i, \bar{\mathbf{v}}_i}
	~&~~\bar{\mathbf{u}}_i^{\top} \mathbf{C} \bar{\mathbf{v}}_i\label{eq:pmc}\\
	{\rm s.\,to }~&~~\bar{\mathbf{u}}_i^{\top}\bar{\mathbf{u}}_i=1,~\bar{\mathbf{v}}_i^{\top}\bar{\mathbf{v}}_i=1\label{eq:pmc2}\\
	~&~~\bar{\mathbf{u}}_i^{\top}\bar{\mathbf{u}}_j=0,~\bar{\mathbf{v}}_i^{\top}\bar{\mathbf{v}}_j=0\label{eq:pmc4}
	\end{align}
\end{subequations}
for all $j=1,\,2,\,\ldots,\, i-1$. 

Since $\mathbf{C}^{\top}\mathbf{C}$ is symmetric, there exists orthonormal matrix $\mathbf{Z}^\ast\in\mathbb{R}^{d\times d}$ and diagonal matrix $\bm{\Lambda}\in\mathbb{R}^{d\times d}$ with diagonal entries $\lambda_1^2\ge \lambda_2^2\ge  \cdots \ge \lambda_d^2$, such that 
\begin{equation}
\label{eq:cc}
(\mathbf{Z}^\ast)^{\top}\mathbf{C}^{\top}\mathbf{C}\mathbf{Z}^\ast= \bm{\Lambda}.
\end{equation} 
 The  columns of $\mathbf{CZ^\ast}$ are orthogonal, and have lengths equal to $\{\lambda_i\ge 0\}$. Concretely, with ${\bf z}_i^*$ denoting the $i$th column of ${\bf Z}^*$, it holds that 
\begin{align}
\label{eq:cz}
(\mathbf{Cz}^\ast_i)^{\top}(\mathbf{Cz}^\ast_j)=
\left\{\begin{array}{cll}
\lambda_i^2,&j = i\\
0,& j\ne i.
\end{array}
\right.
\end{align}

It follows readily that $\mathbf{Z}^\ast:=[\mathbf{z}_1^\ast~\cdots~\mathbf{z}_d^\ast]$ is the optimizer of the following maximization problem 
	\begin{align*}
	\max_{\mathbf{Z}} ~&~~\rm {Tr}(\mathbf {Z}^{\top}\mathbf{C}^{\top}\mathbf{C}\mathbf{Z})\\
	{\rm s.\, to }~&~~\mathbf{Z}^{\top}\mathbf{Z}=\mathbf{I}
	\end{align*} 
which can be equivalently decomposed into $d$ subproblems; that is,
\begin{subequations}
	\label{eq:pca}
	\begin{align}
	\max_{\mathbf{z}_i}~&~~\mathbf{z}_i^{\top}\mathbf{C}^{\top}\mathbf{C}\mathbf{z}_i\label{eq:pcac}\\
	{\rm s. \,to}~&~~\mathbf{z}_i^{\top}\mathbf{z}_i=1,\quad{\rm and}\quad\mathbf{z}_i^{\top}\mathbf{z}_j=0
	\end{align}
\end{subequations}
for $j=1,\,2,\,\ldots,\, i-1$, and  $i=1,\,2\,\ldots,\,d$.

Now we focus on obtaining the first pair of canonical vectors by solving \eqref{eq:pm}. 
After fixing $\bar{\mathbf{v}}_1$, the maximum value of $\bar{\mathbf{u}}_1^{\top}\mathbf{C}\bar{\mathbf{v}}_1$ over $\bar{\mathbf{u}}_1$ is obtained when $\bar{\mathbf{u}}_1$ is proportional to $\mathbf{C}\bar{\mathbf{v}}_1$, meaning
\begin{align}
\label{eq:acbu}
\bar{\mathbf{u}}_1^{\top} \mathbf{C} \bar{\mathbf{v}}_1 \le \|\mathbf{C}\bar{\mathbf{v}}_1\|_2
\end{align}
where the equality is achieved when $\bar{\mathbf{u}}_1={\mathbf{C}\bar{\mathbf{v}}_1}/{\|\mathbf{C}\bar{\mathbf{v}}_1\|_2}$.  
It is clear from \eqref{eq:pca} that $\mathbf{z}_1^\ast$ maximizes $\mathbf{z}_1^{\top}\mathbf{C}^{\top}\mathbf{C}\mathbf{z}_1$.
Thus, $\mathbf{z}_1^\ast$ also maximizes $\|\mathbf{C}\mathbf{z}_1\|_2$, and the maximum of $\|\mathbf{C}\bar{\mathbf{v}}_1\|_2$ is attained when $\bar{\mathbf{v}}_1=\mathbf{z}_1^\ast$, yielding
\begin{align}
\label{eq:acb2}
\bar{\mathbf{u}}_1^{\top}\mathbf{C}\bar{\mathbf{v}}_1 \le \|\mathbf{C}\bar{\mathbf{v}}_1\|_2 \le \|\mathbf{C}\mathbf{z}_1^\ast\|_2 = {\lambda_1}.
\end{align}
When $\bar{\mathbf{u}}_1 = {\mathbf{Cz}_1^\ast}/{\|\mathbf{Cz}_1^\ast \|_2}$ and $\bar{\mathbf{v}}_1 = \mathbf{z}_1^\ast$, the first two inequalities in \eqref{eq:acb2} hold as equalities, proving that $\bar{\mathbf{u}}_1^{\ast} = {\mathbf{Cz}^\ast_1}/{\|\mathbf{Cz}^\ast_1\|_2}$ and $\bar{\mathbf{v}}_1^{\ast} = \mathbf{z}^\ast_1$. 

After finding the optimal $\bar{\mathbf{u}}_1^{\ast}$ and $\bar{\mathbf{v}}_1^{\ast}$, one can further search for $\bar{\mathbf{u}}_i^\ast$ and $\bar{\mathbf{v}}_i^\ast$ for $i\ge 2$ by solving \eqref{eq:pm}. 
Without considering the constraints in \eqref{eq:pmc4}, we find  $\bar{\mathbf{v}}_i^{\ast}=\mathbf{z}^\ast_i$ and $\bar{\mathbf{u}}_i^{\ast}={\mathbf{Cz}^\ast_i}/{\|\mathbf{Cz}^\ast_i\|_2}$, which can be proved in the same way argued for $i=1$. 
Next, we show that $\bar{\mathbf{v}}_i^{\ast}$ and $\bar{\mathbf{u}}_i^{\ast}$ satisfy constraints \eqref{eq:pmc4}. Obviously, $\bar{\mathbf{v}}_i^{\ast}$ is orthogonal to all vectors in the set  $\{\bar{\mathbf{v}}^{\ast}_j\}_{j=1}^{i-1}$. Furthermore, $(\bar{\mathbf{u}}_i^{\ast})^{\top}\bar{\mathbf{u}}_j^{\ast}=(\mathbf{Cz}^\ast_i)^{\top}\mathbf{Cz}^\ast_j/(\|\mathbf{Cz}^\ast_i\|_2\|\mathbf{Cz}^\ast_j\|_2)$, and \eqref{eq:cz} implies  that $\bar{\mathbf{u}}_i^{\ast}$ is orthogonal to $\bar{\mathbf{u}}^{\ast}_j$ for $j=1,\,2,\,\ldots,\,i-1$.

Summarizing the two cases, we deduce that $\bar{\mathbf{u}}_i^{\ast}={\mathbf{Cz}^\ast_i}/{\|\mathbf{Cz}^\ast_i\|_2}$ and $\bar{\mathbf{v}}_i^{\ast}=\mathbf{z}^\ast_i$ for $i=1,\,2,\,\ldots,\, d$, and $\bar{\mathbf{u}}_i^{\ast}$ and $\bar{\mathbf{v}}_i^{\ast}$ are the $i$-th left and right singular vector of $\mathbf{C}$ associated with the $i$-th largest singular value, that is ${\lambda}_i$.

In general, there may be zero eigenvalues. Suppose that the last positive eigenvalue is $\lambda_k^2$; in other words, it holds that $\lambda_1^2
\ge \,\cdots\, \ge \lambda_k ^2> \lambda_{k+1}^2=\,\cdots\, =
\lambda_d^2=0$. As such, the optimal solutions $\{\bar{\mathbf{u}}_i^\ast, \,\bar{\mathbf{v}}_i^\ast\}_{i=k+1}^{d}$ can be any set of $d$ vectors satisfying constraints \eqref{eq:pmc2} and \eqref{eq:pmc4}.  
 
Once having computed $\bar{\mathbf{U}}^{\ast}=[\bar{\mathbf{u}}_1^{\ast}~\cdots~\bar{\mathbf{u}}_d^{\ast}]$ and $\bar{\mathbf{V}}^{\ast}=[\bar{\mathbf{v}}_1^{\ast}~\cdots~\bar{\mathbf{v}}_d^{\ast}]$, the optimal solutions $\mathbf{U}^\ast$ and $\mathbf{V}^\ast$ to problem \eqref{eq:gpcca} are obtained as $\mathbf{U}^{\ast}:=\boldsymbol{\Sigma}_x^{-1/2}\bar{\mathbf{U}}^{\ast}$ and $\mathbf{V}^{\ast}:=\boldsymbol{\Sigma}_y^{-1/2}\bar{\mathbf{V}}^{\ast}$. Moreover, the maximal value of \eqref{eq:gpccac} becomes $\sum_{i=1}^{d}{\lambda_i}$.

\subsection{Proof of Theorem \ref{th:gkcca}}\label{profgkcca}
Upon defining 
\begin{align*}
&\bar{\mathbf{A}}:=(\mathbf{K_x+\epsilon\mathbf{I}})^{1/2}\mathbf{K}_x^{1/2}\mathbf{A} \\
&\bar{\mathbf{B}}:=(\mathbf{K_y+\epsilon\mathbf{I}})^{1/2}\mathbf{K}_y^{1/2}\mathbf{B}
\end{align*} 
problem \eqref{eq:gkccam} can be rewritten as
\begin{subequations}
	\label{eq:gkccam_32}
	\begin{align}
(\bar{\mathbf{A}}^{\ast},\, \bar{\mathbf{B}}^{\ast}) := \arg\max_{\bar{\mathbf{A}}, \, \bar{\mathbf{B}}} ~&~{\rm{Tr}}(\bar{\mathbf{A}}^{\top}\mathbf{C}\bar{\mathbf{B}})\label{eq:gkccam_3c20}\\
	{\rm s. \,to } ~&~\bar{\mathbf{A}}^{\top}\bar{\mathbf{A}}=\mathbf{I} ,~ {\rm and}~\bar{\mathbf{B}}^{\top}\bar{\mathbf{B}}=\mathbf{I} .\label{eq:gkccam_3c22}
	\end{align} 
\end{subequations} 
Using the results in Appendix \ref{prof2}, one readily concludes that 
 the columns of optimizers $\bar{\mathbf{A}}^{\ast}$, $\bar{\mathbf{B}}^{\ast}$  consist of the $d$ left and right singular vectors of $\mathbf{C}$ associated with the first $d$ largest singular values, respectively, which leads  to
\begin{align*} 
\mathbf{A}^{\ast}&=\mathbf{K}_x^{-1/2}(\mathbf{K}_x+\epsilon \mathbf{I})^{-1/2}\bar{\mathbf{A}}^{\ast}
\\
\mathbf{B}^{\ast}&=\mathbf{K}_y^{-1/2}(\mathbf{K}_y+\epsilon \mathbf{I})^{-1/2}\bar{\mathbf{B}}^{\ast}.
\end{align*}
Likewise, the maximal value of \eqref{eq:gkccamc} is given by the sum  of the $d$ largest singular values of $\mathbf{C}$.

\bibliographystyle{IEEEtran}
\bibliography{pca}

\end{document}